%% file: main.tex
\documentclass[10pt,journal,compsoc]{IEEEtran}

\ifCLASSOPTIONcompsoc
  \usepackage[nocompress]{cite}
\else
  \usepackage{cite}
\fi

\usepackage{bm}
\usepackage{amsmath}
\usepackage{amssymb}
\usepackage{array}
\usepackage[algo2e,inoutnumbered,linesnumbered,algoruled,vlined]{algorithm2e}
\usepackage{algpseudocode}
\usepackage{graphicx}
\usepackage{booktabs}
\usepackage{multirow}
\usepackage{xcolor}
\usepackage[subfigure]{tocloft}
\usepackage[pagebackref=true,breaklinks=true,colorlinks,bookmarks=false]{hyperref}

\ifCLASSOPTIONcompsoc
 \usepackage[caption=false,font=footnotesize,labelfont=sf,textfont=sf]{subfig}
\else
 \usepackage[caption=false,font=footnotesize]{subfig}
\fi

\usepackage{url}
\input{defs}

\begin{document}

\setlength{\abovedisplayskip}{2pt}
\setlength{\belowdisplayskip}{2pt}

\title{Multiway Non-rigid Point Cloud Registration via Learned Functional Map Synchronization}

\author{Jiahui Huang,
        Tolga Birdal, Zan Gojcic, \\
        Leonidas J. Guibas~\IEEEmembership{Senior Member,~IEEE}
        and~Shi-Min Hu~\IEEEmembership{Senior Member,~IEEE}%
        \IEEEcompsocitemizethanks{\IEEEcompsocthanksitem J. Huang and S.M. Hu are with the BNRist, Department of Computer Science and Technology, Tsinghua University.\protect\\
        Shi-Min Hu is the corresponding author.
        \IEEEcompsocthanksitem T. Birdal and L. J. Guibas are with Stanford University.
        \IEEEcompsocthanksitem Z. Gojcic is with Nvidia.}}

\markboth{Transactions on Pattern Analysis and Machine Intelligence, arxiv preprint}
{Huang \MakeLowercase{\textit{et al.}}: Multiway Non-rigid Registration.}

\IEEEtitleabstractindextext{%
\input{sections/grp0-abstract}

}

\maketitle

\IEEEdisplaynontitleabstractindextext
\IEEEpeerreviewmaketitle

\input{sections/grp1-intro}

\input{sections/grp2-related}

\input{sections/grp3-method}

\input{sections/grp4-exp}

\input{sections/grp5-ending}



\ifCLASSOPTIONcompsoc
  \section*{Acknowledgments}
\else
  \section*{Acknowledgment}
\fi
This paper was supported by National Key R\&D Program of China (project No. 2021ZD0112902), a Vannevar Bush faculty fellowship, ARL grant W911NF2120104, NSF grant IIS-1763268, and a gift from the Autodesk Corporation.

\ifCLASSOPTIONcaptionsoff
  \newpage
\fi

\bibliographystyle{IEEEtran}
\bibliography{IEEEabrv,reference}

\newpage

\vspace{-3em}

\begin{IEEEbiography}[{\includegraphics[width=1in,height=1.25in,clip,keepaspectratio]{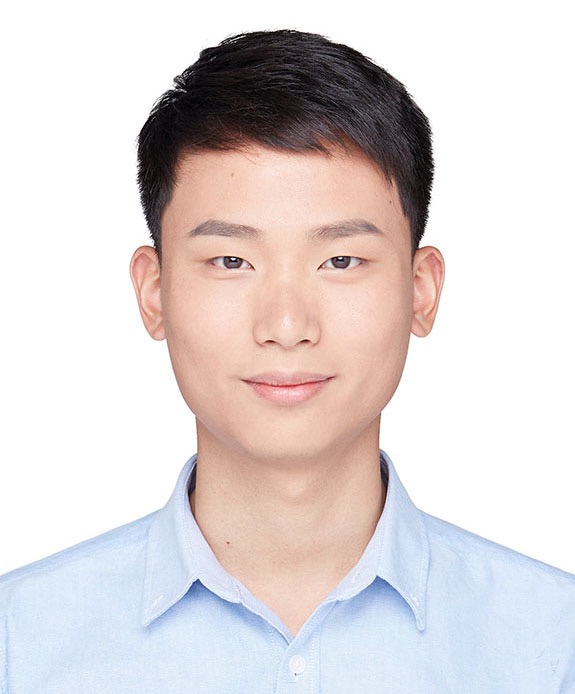}}]{Jiahui Huang}
received his B.S. degree in computer science and technology from Tsinghua University in 2018. 
He is currently a Ph.D. candidate at the Departemnt of Computer Science and Technology, Tsinghua University. 
His research interests include computer vision, robotics and computer graphics.
His personal webpage is at \url{https://cg.cs.tsinghua.edu.cn/people/~huangjh/}.
\end{IEEEbiography}

\vspace{-3em}

\begin{IEEEbiography}[{\includegraphics[width=1in,height=1.35in,clip,keepaspectratio,trim=20 0 20 0]{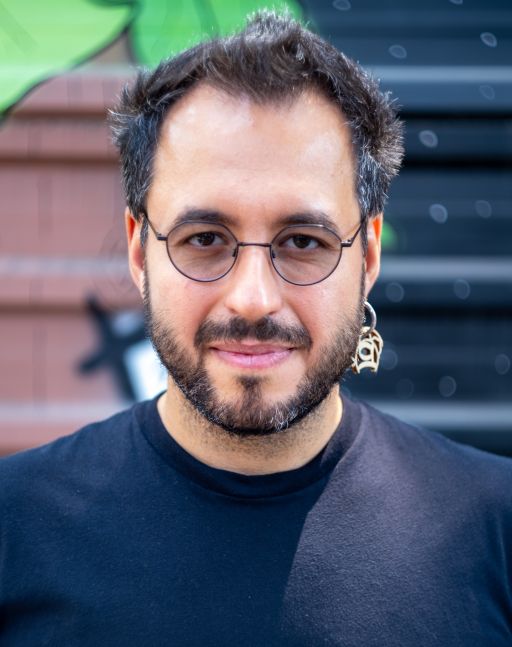}}]{Tolga Birdal}
is a postdoctoral researcher at the Geometric Computing Group of Stanford University. He has defended his masters and Ph.D. theses at the Computer Vision Group, CAMP Chair, TU Munich. He was also a Doktorand at Siemens AG. His current foci of interest involve geometric machine learning and 3D computer vision. More theoretical work is aimed at investigating the limits in geometric computing, non-Euclidean inference and principles of deep learning.
His personal webpage is at \url{https://www.tbirdal.me}.
\end{IEEEbiography}

\vspace{-3em}

\begin{IEEEbiography}[{\includegraphics[width=1in,height=1.25in,clip,keepaspectratio,trim=10 0 10 0]{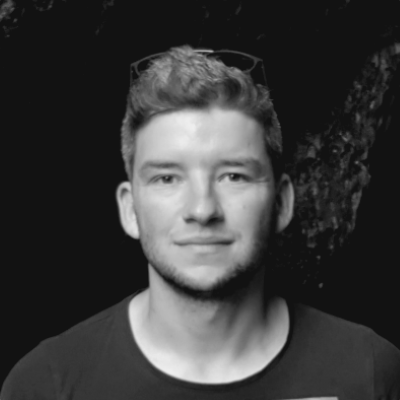}}]{Zan Gojcic} is currently a research scientist at the NVIDIA AI Lab, which he joined in 2021. He has received his PhD degree from ETH Zurich in 2021, under the supervision of Andreas Wieser. His research interests revolve around 3D computer vision, domain adaptation, and reducing the supervision. His personal webpage is located at \url{https://zgojcic.github.io/}
\end{IEEEbiography}

\vspace{-3em}

\begin{IEEEbiography}[{\includegraphics[width=1in,height=1.35in,clip,keepaspectratio]{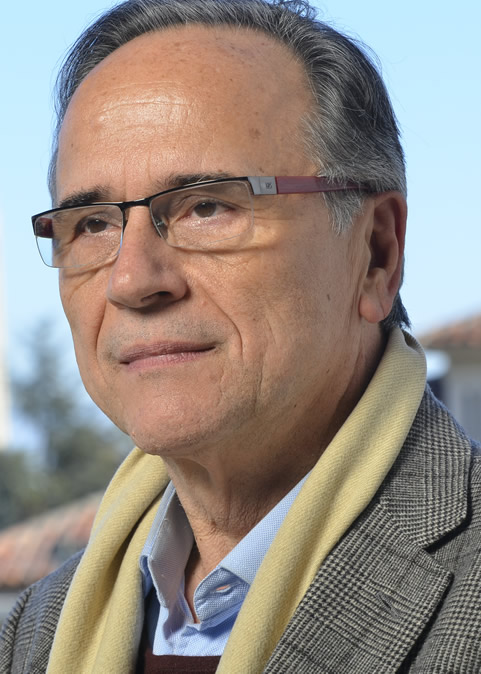}}]{Leonidas J. Guibas}
received his Ph.D. degree from Stanford University in 1976, under the supervision of Donald Knuth. His main subsequent employers were Xerox PARC, MIT, and DEC/SRC. Since 1984, he has been at Stanford University, where he is a professor of computer science. His research interests include computational geometry, geometric modeling, computer graphics, computer vision, sensor networks, robotics, and discrete algorithms. He is a senior member of the IEEE and the IEEE Computer Society. 
\end{IEEEbiography}

\vspace{-3em}

\begin{IEEEbiography}[{\includegraphics[width=1in,height=1.25in,clip,keepaspectratio]{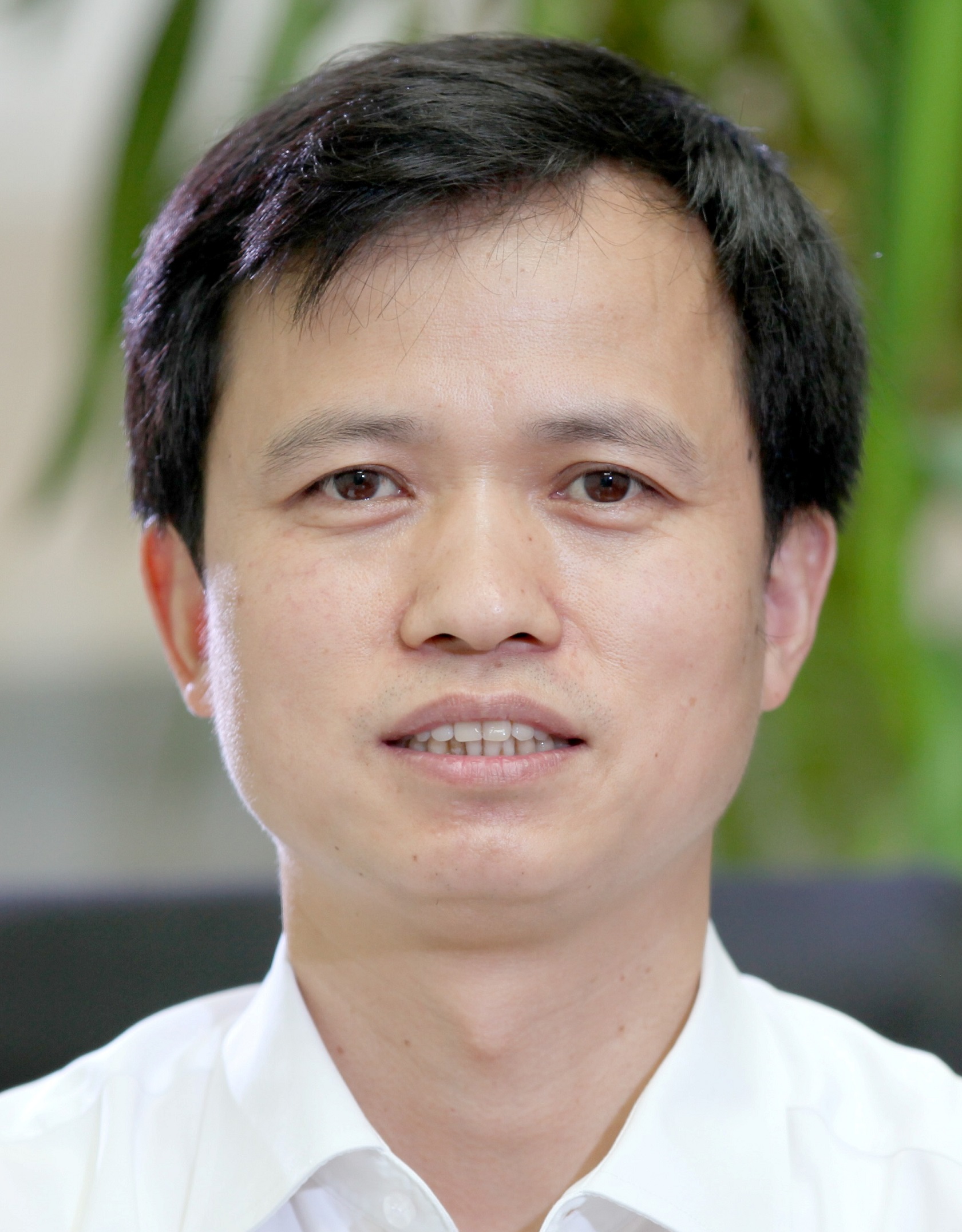}}]{Shi-Min Hu} 
is currently a professor in Computer Science at Tsinghua University.
He received a Ph.D. degree from Zhejiang University in 1996.
His research interests include geometry processing, image \& video processing, rendering, computer animation and CAD.
He has published more than 100 papers in journals and refereed conferences.
He is Editor-in-Chief of Computational Visual Media, and on the editorial boards of several journals, including Computer Aided Design and Computer \& Graphics.
\end{IEEEbiography}


\vfill

\clearpage
\setcounter{equation}{0}
\setcounter{figure}{0}
\setcounter{algocf}{0}

\renewcommand{\thetable}{S\arabic{table}}
\renewcommand{\thefigure}{S\arabic{figure}}
\renewcommand{\theequation}{S.\arabic{equation}}
\makeatletter
\renewcommand{\thealgocf}{S\@arabic\c@algocf}
\makeatother

\appendices
\newcommand\refpaper[1]{\cref{#1}}
\input{sections/grp6-appendix}

\end{document}

%% file: defs.tex
\usepackage{pifont}
\usepackage{tikz}
\usepackage{multirow}
\usepackage{enumitem}
\usepackage{microtype}
\usepackage{wrapfig}
\usepackage[column=0]{cellspace}

\makeatletter
\DeclareRobustCommand\onedot{\futurelet\@let@token\@onedot}
\def\@onedot{\ifx\@let@token.\else.\null\fi\xspace}

\def\eg{\emph{e.g}\onedot} 
\def\ie{\emph{i.e}\onedot} 
\def\cf{\emph{c.f}\onedot} 
 
\def\wrt{w.r.t\onedot} 

\makeatother

\newcommand{\x}{\bm{x}} 
\newcommand{\X}{\mathbf{X}} 
\newcommand{\flow}{\mathbf{F}} 
\newcommand{\Id}{\mathbf{I}} 
\newcommand{\R}{\mathbb{R}} 
\newcommand{\Lap}{\mathbf{L}} 
\newcommand{\basis}{\mathbf{\Phi}} 
\newcommand{\C}{\mathbf{C}} 
\newcommand{\Perm}{\mathbf{\Pi}}
\newcommand{\Hfunc}{\mathbf{H}} 
\newcommand{\h}{\mathbf{h}} 
\newcommand{\desc}{\mathbf{D}}
\newcommand{\hwt}{\mathbf{\Lambda}}     

\newcommand{\A}{\mathbf{A}} 
\newcommand{\B}{\mathbf{B}} 
\newcommand{\bb}{\mathbf{b}} 

\newcommand{\basisnet}{\varphi_{\mathrm{basis}}} 
\newcommand{\descnet}{\varphi_{\mathrm{desc}}} 
\newcommand{\refnet}{\varphi_{\mathrm{refine}}} 
\newcommand{\corr}{\mathfrak{g}}  

\newcommand{\one}{\mathbf{1}} 
\newcommand{\zero}{\mathbf{0}} 

\newcommand{\gt}{\mathrm{gt}}   
\newcommand{\pd}{\mathrm{pd}}   

\newcommand{\Graph}{\mathcal{G}}
\newcommand{\Vertex}{\mathcal{V}}
\newcommand{\Edge}{\mathcal{E}}
\newcommand{\diag}{\mathrm{diag}}
\renewcommand{\vec}{\mathrm{vec}}

\newcommand{\Shape}{\mathcal{S}}
\newcommand{\warp}{\mathcal{W}}
\newcommand{\Fset}{\mathcal{F}}
\newcommand{\loss}{\mathcal{L}}
\newcommand{\Cset}{\mathcal{C}}
\newcommand{\Aset}{\mathcal{A}}
\newcommand{\human}{\mathrm{H}}
\newcommand{\knn}{\mathcal{N}}

\newcommand{\de}{\mathrm{d}}
\newcommand{\nn}{\mathrm{n}}      
\newcommand{\bs}{\mathrm{f}}      
\newcommand{\fro}{\mathrm{F}}      

\newcommand{\name}{{SyNoRiM\xspace}}

\newcommand{\dcape}{{MPC-CAPE\xspace}}
\newcommand{\ddd}{{MPC-DD\xspace}}
\newcommand{\ddtfd}{{MPC-DT4D\xspace}}
\newcommand{\dsapien}{{MPC-SAPIEN\xspace}}
\newcommand{\epe}{{L2 Error\xspace}}
\newcommand{\jetbar}{\includegraphics[width=3em,height=0.5em]{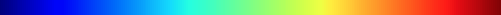}}
\newcommand{\wosync}{{\scriptsize(\textls[-120]{ w/o } sync.)}}

\DeclareMathOperator*{\argmin}{arg\min}

\newcommand{\refap}[1]{{\color{red}#1}}

\usepackage{cleveref}
\crefname{section}{\S}{\S\S}
\crefname{subsection}{\S}{\S\S}
\crefname{conj}{Conj.}{Conj.}

\Crefname{assumption}{\textbf{H}\hspace{-3pt}}{\textbf{H}\hspace{-3pt}}
\crefname{assumption}{\textbf{H}}{\textbf{H}}

\crefname{algorithm}{\text{Alg.}}{\text{Alg.}}
\crefname{assumption}{\textbf{H}}{\textbf{H}}
\crefname{equation}{\text{Eq}}{\text{Eq}}
\crefname{definition}{\text{Dfn.}}{\text{Dfn.}}
\crefname{lemma}{\text{Lemma}}{\text{Lemma}}
\crefname{dfn}{\text{Dfn.}}{\text{Dfn.}}
\crefname{thm}{\text{Thm.}}{\text{Thm.}}
\crefname{tab}{\text{Tab.}}{\text{Tab.}}
\crefname{fig}{\text{Fig.}}{\text{Fig.}}
\crefname{table}{\text{Tab.}}{\text{Tab.}}
\crefname{figure}{\text{Fig.}}{\text{Fig.}}

\newcommand{\parahead}[1]{\vspace{1.5mm}\noindent\textbf{#1.}\ }

\definecolor{caribbeangreen}{rgb}{0.0, 0.8, 0.6}
\definecolor{amethyst}{rgb}{0.6, 0.4, 0.8}
\definecolor{fandango}{rgb}{0.91, 0.3, 0.64}
\definecolor{set3_green}{rgb}{0.56, 0.83, 0.78}
\definecolor{set3_yellow}{rgb}{1.0, 1.0, 0.72}
\definecolor{set3_purple}{rgb}{0.75, 0.73, 0.85}
\definecolor{set3_red}{rgb}{0.98, 0.51, 0.46}
\DeclareRobustCommand{\cbox}[1]{\tikz \fill[#1] (0,0) rectangle (1ex, 1ex);}

\newcommand{\listnumberedparaheadname}{List of Numbered Paraheads}
\newlistof{numberedparahead}{ex}{\listnumberedparaheadname}
\renewcommand\thenumberedparahead{\Alph{numberedparahead}}
\newcommand{\numberedparahead}[1]{%
\refstepcounter{numberedparahead}%
\parahead{\thenumberedparahead. #1}}
\makeatletter
\@addtoreset{numberedparahead}{subsection}
\makeatother
\crefname{numberedparahead}{\S}{\S\S}

\definecolor{reviewer_box}{rgb}{0.9, 0.95, 1.0}
\definecolor{reviewer_question}{rgb}{0.3, 0.35, 0.8}

\newcommand{\rev}[1]{#1} 

\newif\ifresponse

%% file: sections/grp0-abstract.tex
\begin{abstract}
\rev{We present \name, a novel way to jointly register multiple non-rigid shapes by synchronizing the maps that relate learned functions defined on the point clouds.}
Even though the ability to process non-rigid shapes is critical in various applications ranging from computer animation to 3D digitization, the literature still lacks a robust and flexible framework to match and align a collection of real, noisy scans observed under occlusions. 
Given a set of such point clouds, our method first computes the pairwise correspondences parameterized via functional maps. We simultaneously learn potentially non-orthogonal basis functions to effectively regularize the deformations, while handling the occlusions in an elegant way. 
To maximally benefit from the multi-way information provided by the inferred pairwise deformation fields, we synchronize the pairwise functional maps into a \textit{cycle-consistent} whole thanks to our novel and principled optimization formulation. 
We demonstrate via extensive experiments that our method achieves a state-of-the-art performance in registration accuracy, while being flexible and efficient as we handle both non-rigid and multi-body cases in a unified framework and avoid the costly optimization over point-wise permutations by the use of basis function maps.
Our code is available at \url{https://github.com/huangjh-pub/synorim}.
\end{abstract}

%% file: sections/grp1-intro.tex
\IEEEraisesectionheading{\section{Introduction}
\label{sec:introduction}}
\IEEEPARstart{T}{he} prevalence of reliable 3D data capture fueled countless applications impacting from movie industry to robotics. 
In a wide variety of these applications, one needs to capture (in 3D) non-rigidly moving objects from multiple angles or over time~\cite{dou2013scanning,Collet2015HighqualitySF}. 
This leads to a dynamic, multi-scan alignment problem, further obstructed by the presence of occlusions, ambiguities, and noise.

Solving this challenging task fostered the development of a plethora of temporal, mesh-based dynamic non-rigid registration algorithms (\eg~\cite{newcombe2015dynamicfusion}). 
However, these methods suffer from two main limitations: 
(1) the data provided by 3D sensors hardly come in mesh format, let alone the difficulty associated with preserving the mesh topology.
Signed distance field based reconstruction methods like~\cite{slavcheva2018sobolevfusion} can overcome some of these problems, but (2) they still assume a \emph{streaming} depth map and fail to maintain correspondences, which are critical for defining non-rigid deformations. 
In addition to those nuisances, a drift-free alignment almost surely demands a global optimization step, which exploits all possible loop closure constraints in the form of a graph optimization~\cite{birdalSimsekli2018} or bundle adjustment~\cite{triggs1999bundle}.

\begin{figure}[!t]
\centering
\vspace{-2em}
\includegraphics[width=\linewidth]{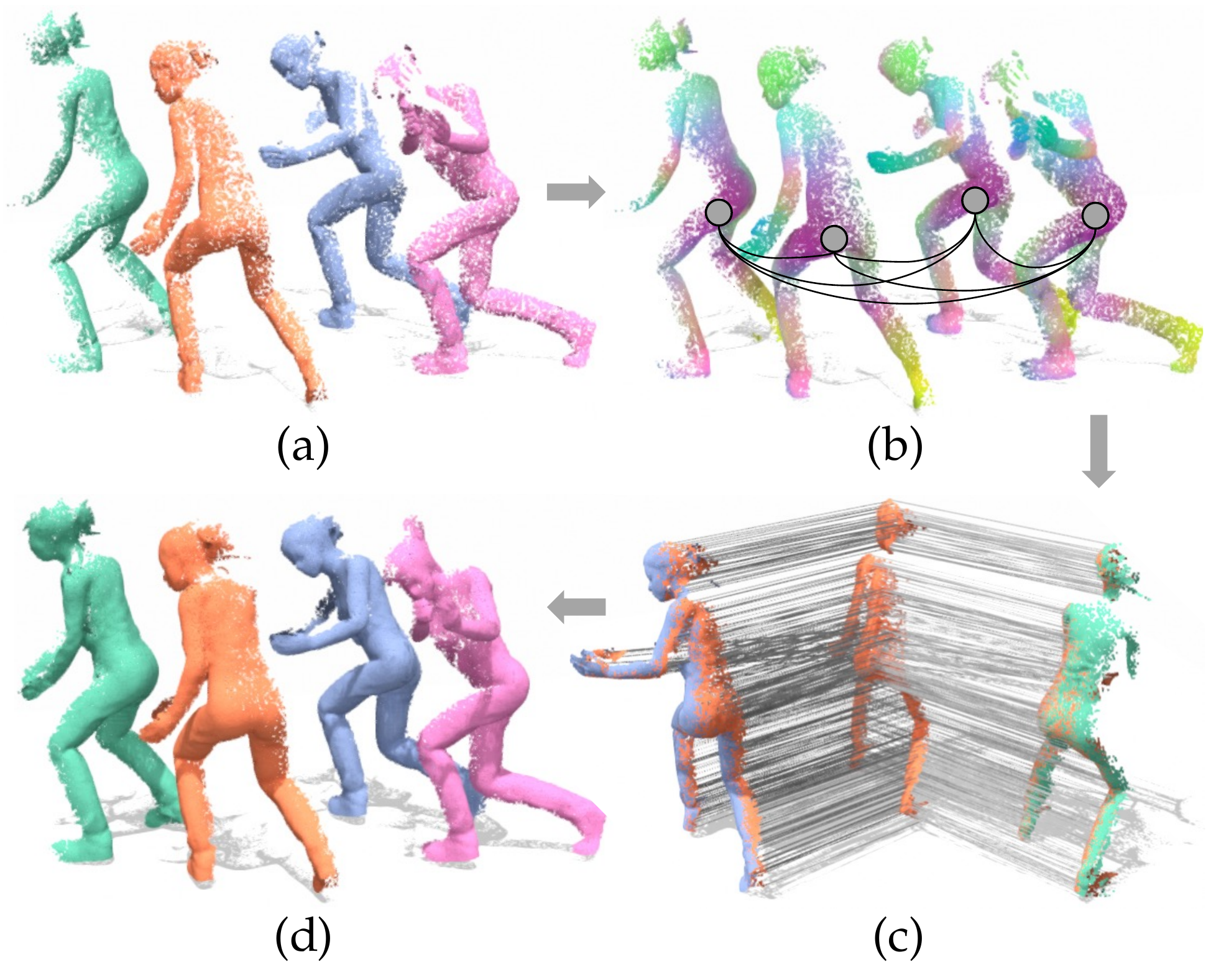}
\vspace{-2em}
\caption{\textbf{\name~overview}. (a) Input point clouds. (b) Synchronized canonical functions $\Hfunc$ (visualized via Principal Component Analysis; see \cref{sec:sync} for details). (c) Scene flow estimations for all pairs (\rev{here only two sets of flow, \cbox{orange}$\rightarrow$\cbox{caribbeangreen} and \cbox{orange}$\rightarrow$\cbox{amethyst}, are shown for brevity}). (d) Registered (accumulated) point clouds by gathering warped points from other inputs.}
\vspace{-2em}
\label{fig:teaser}
\end{figure}
In this paper, we set off in pursuit of alleviating both of these issues. In particular, we (1) leverage unstructured point cloud representations for maximal generality and flexibility; (2) assume a fully connected graph in lieu of the sequential order, incorporating drift reduction in the early stages. 
Additionally, \rev{compared to rigid motion~\cite{huang2021multibodysync,Gojcic2021WeaklySL} or pure as-rigid-as-possible~\cite{SorkineHornung2007AsrigidaspossibleSM} deformation, our relaxed assumptions allow for more general object dynamics.}
%
In particular, we propose \textbf{\name}~framework, \underline{\textbf{sy}}nchronization for \underline{\textbf{no}}n-\underline{\textbf{ri}}gid \underline{\textbf{m}}ultiway registration.
Input to \name, is a collection of point cloud scans containing a potentially non-rigidly deforming object. We then seek to recover a coherent and consistent 3D scene flow, originating from any source shape to a latent target shape that is to be discovered simultaneously. We start by a \textit{motion coherence observation}~\cite{yuille1988motion} that pairwise correspondences between natural shapes are smooth and band-limited, \ie nearby points in the source point cloud should map to nearby points in the target (except for topological changes and occlusions). This allows us to substitute the point-wise matches between pairs of scans by linear maps of \emph{learned smooth functions} defined on the points. Such notion of a \textit{functional map} was first introduced in the geometry processing community~\cite{ovsjanikov2012functional} for estimating correspondences between isometric meshes. Unfortunately, due to the lack of proper basis functions, an extension to point cloud analysis is not trivial~\cite{huang2018stability}. In \name~we propose to directly learn the bases from a large-scale training corpus, either in a fully-supervised or unsupervised fashion. Such bases learned by neural networks are found to be robust to occlusions, partiality, and errors in the initialization because an end-to-end training informs the bases about the shape context while the inherent smoothness enables meaningful extrapolations of the target coordinate functions. 
A subsequent \textit{functional map synchronization} algorithm coerces the maps to be globally consistent, by making use of the information that individual observations are essentially samples from a common underlying surface. Based on this, we finally refine the pairwise deformations eliminating the ambiguities. An illustrative figure is shown in \cref{fig:teaser}.

In summary, our contributions are:
\begin{enumerate}[leftmargin=*]
	\item We propose, to the best of our knowledge, the first end-to-end data-driven framework to learn consistent registration between multiple, possibly partial point cloud observations of non-rigidly moving bodies / objects.
	\item Our deep network can learn potentially non-orthogonal functional bases on point sets without requiring to define Laplace-Beltrami operators (LBO). Functional maps between such bases can match non-isometric shapes.
	\item We propose a novel \textit{functional map synchronization} algorithm enforcing cycle-consistency among the pairwise deformation fields, estimated in isolation. This harmonizes the 3D flows into a coherent whole and thus enables multiway registration.
\end{enumerate}

We demonstrate the efficacy of our algorithm through extensive evaluations on both rigid and non-rigid scenarios, showing superior performance on all datasets.

%% file: sections/grp2-related.tex
\section{Related Works}
\label{sec:related}

\parahead{Point-set Registration}
Finding reliable alignments between point clouds plays a fundamental role for many downstream tasks.
Rigid registration estimates a single transformation matrix through either heuristic searches~\cite{icp} or learned local/global descriptors~\cite{ao2021spinnet,huang2021predator,zhao2020quaternion,deng2018ppfnet,deng20193d}, while common non-rigid registration techniques aim to best align the clouds under the various data terms~\cite{amberg2007optimal,myronenko2010cpd,yao2020fastrnrr} and regularizations~\cite{SorkineHornung2007AsrigidaspossibleSM,slavcheva2018sobolevfusion} with different deformation representations~\cite{sumner2007embedded,bozic2021neural}.
In this paper we consider the general non-rigid deformation scheme and also demonstrate results on hybrid ones such as multibody~\cite{costeira1998multibody,huang2019clusterslam,huang2020clustervo}.
On the other hand, scene flow describes the transition between two point clouds using a three-dimensional vector field and is a low-level task agnostic of the deformation type.
Existing techniques~\cite{baur2021slim,wu2019pointpwc,puy20flot,li2021hcrf} handle the task via accurate modeling of the point-wise features as well as its neighborhood context, and reach a good performance for cluttered scenes or driving scenarios, but is not robust under large deformations and ambiguities~\cite{ouyang2021occlusion}.

\parahead{Function-based Correspondence}
In the field of geometry processing, the use of functional techniques is a recent trend for building reliable correspondences between 3D shapes (\eg, discretized manifold meshes).
First introduced in \cite{ovsjanikov2012functional}, such methods compute the LBO eigenvectors as basis functions and infer a linear transformation of a subset of bases to indicate low-rank shape mapping.
Many extensions have been vastly explored such as hierarchical matching~\cite{melzi2019zoomout}, partial-to-full handling~\cite{litany2017fully} or integration into deep learning frameworks~\cite{donati2020deep,litany2017deep} with either supervised or unsupervised~\cite{eisenberger2020deep,roufosse2019unsupervised,halimi2019unsupervised} methods.
Contrarily in the domain of point set analysis, such methods are less popular despite a few~\cite{magnet2021dwks,marin2020correspondence,Sharp2020DiffusionNetDA}. 
\rev{Notably, \cite{marin2020correspondence} also propose to learn a linearly-invariant embedding as bases, yet their learning scheme is different from ours and they are not robust to partialities or occlusions, which are commonly observed in point cloud data.}
We further highlight that the idea to project natural signals to lower dimensions has also been partially explored in the vision community~\cite{tang2020lsm,ye2021motion}.

\parahead{Synchronization on 3D Geometry}
Though initially established as a theory for clock systems, synchronization has now been widely used in many vision tasks such as structure from motion~\cite{hartley2013rotation,birdal2020synchronizing,birdalSimsekli2018}, semantic segmentation~\cite{wang2013image}, \rev{correspondence refinement}~\cite{birdal2019probabilistic,birdal2021quantum} and multiway rigid/multibody registration~\cite{huang2020uncertainty,gojcic2020learning,huang2021multibodysync}.
By enforcing cycle consistencies within a system, the relative measurements are globally harmonized thanks to the averaging of local noise.
In the field of 3D geometry analysis, analyzing and synchronizing spectral functional maps~\cite{zhang2010spectral} from a mesh collection are widely applied to full shape matching, either with low-rank factorization~\cite{huang2014functional}, limit shapes extraction~\cite{huang2019limit}, coarse-to-fine strategy~\cite{huang2020consistent} or joint point-spectrum optimization~\cite{gao2021isometric}.
In contrast, our novel synchronization formulation is tailored for the bases learned from raw point clouds that are free of the geometric impositions like (near-)isometries and directly takes correspondences into account, robustly.

%% file: sections/grp3-method.tex
\section{Overview}
\label{sec:overview}

\parahead{Problem Formulation}
The input to our method is a point cloud graph $\Graph := (\Vertex, \Edge)$, whose vertices $\Vertex$ represent the input set of $K$ point clouds $\Vertex := \{\X_k \in \R^{N_k \times 3}, k \in [1,K] \}$ and the edges $\Edge := \{(k,l), \X_k \in \Vertex, \X_l \in \Vertex\}$ represent the graph connectivity. By default we assume $\Graph$ to be fully-connected.
The output of our method contains all the pairwise per-point 3D flow vectors $\Fset := \{ \flow_{kl} \in \R^{N_k \times 3}, (k,l) \in \Edge \}$.
The flow vectors naturally induce the non-rigid warp field from $\X_k$ to $\X_l$ as $\warp_{kl} (\X_k) := \X_k + \flow_{kl}$, optimally aligning the given point cloud pairs by deforming the source onto the target.
We additionally encourage the cyclic consistency of the estimates $\Fset$, defined loosely as:
\begin{equation}
    \warp_{k_1 k_2} \circ \dots \circ \warp_{k_{p-1} k_p} \circ \warp_{k_p k_1} = \Id, \forall (k_1, \dots k_p)\in \mathcal{C}(\Graph),
\end{equation}
where $\mathcal{C}(\Graph)$ are the set of cycles in the $\Graph$ and $\Id$ is the identity warping.
The domain of the above composed warp map is the union region $(\X_{k_1}, \dots, \X_{k_p})$.

\begin{figure*}[!t]
\centering
\includegraphics[width=\linewidth]{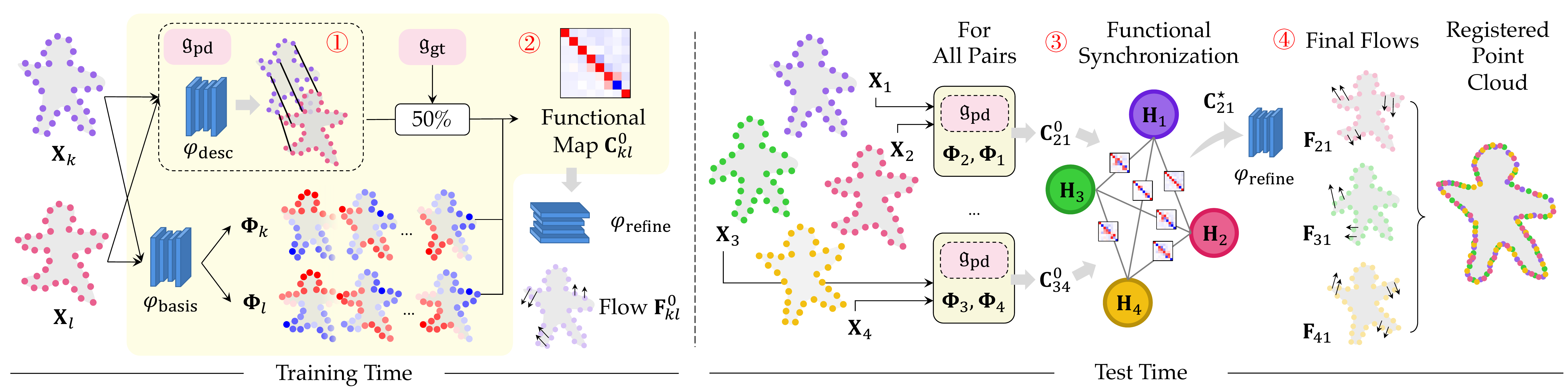}
\vspace{-2.5em}
\caption{\rev{\textbf{Method.} During training, our method is supervised in a pairwise fashion. We first train \ding{172} $\descnet$ (\ie, $\corr_\pd$) to establish putative correspondences between each point cloud pair. We then estimate a set of basis functions $\{\basis_k\}$ for each point cloud using $\basisnet$ to obtain \ding{173} the initial functional map $\C_{kl}^0$ before refining the 3D flow estimates with $\refnet$. During test time with multiple inputs, we estimate the map set $\{\C_{kl}^0\}$ for all pairs. \ding{174} The estimated maps are subsequently synchronized to optimize cycle consistency among the inputs. Finally, \ding{175} 3D flows are estimated from the optimized functional maps $\{\C_{kl}^\star\}$ as our final output. The registered point cloud is a fusion of all initial point clouds warped by the estimated flows.}}
\label{fig:pipeline}
\vspace{-1em}
\end{figure*}

\parahead{Method Summary}
Our method begins by establishing sparse point-level putative correspondences using a correspondence generator $\corr(\cdot)$ (\cref{subsec:corr}).
In the meantime, a basis network $\basisnet$ is applied independently to each input point cloud to generate a set of basis functions defined on the points (\cref{subsec:basis}).
Given the sparse correspondences and bases we can compute the initial pairwise functional map matrices.
The technique to recover flows from the matrices is realized through another network $\refnet$ (\cref{subsec:sfffm}).
We then feed the initial map estimates into a synchronization module that jointly optimizes for all pairwise functional mappings, considering the cycle consistency (\cref{sec:sync}), before utilizing the $\refnet$ module once more to get the optimized deformations.
The full pipeline is illustrated in~\cref{fig:pipeline}.

\section{Pairwise Functional Registration}
\label{sec:pw}

\subsection{Preliminaries on Functional Maps}
\label{subsec:fmap_basic}

Given two (abstract) shapes $\Shape_k$ and $\Shape_l$ in the form of smooth continuous manifolds, a functional map $T_{kl}: L_2(\Shape_l) \mapsto L_2(\Shape_k)$ maps from the space of square-integrable real-valued \emph{functions} ($L_2$-space) defined on $\Shape_l$ to $\Shape_k$.
Such an operator is proved to be linear~\cite{ovsjanikov2012functional}, \ie, $T_{kl} (\alpha_1 f_1 + \alpha_2 f_2) = \alpha_1 T_{kl} (f_1) + \alpha_2 T_{kl} (f_2)$, where $f_1$ and $f_2$ are functions defined on $\Shape_l$ and $\alpha_1$ and $\alpha_2$ are the coefficients.

Every function $f_k$ on $\Shape_k$ (or $f_l$ on $\Shape_l$) can be represented as a linear combination of \emph{basis functions} $\{\phi_{k,m}\}$ (or $\{\phi_{l,m}\}$): $f_k=\sum_m h_{k,m} \phi_{k,m}, f_l = \sum_m h_{l,m} \phi_{l,m}$, with $h_{k,m}$ and $h_{l,m}$ being the coefficients.
Moreover, most natural functions defined on shapes can be approximated by linearly combining a \emph{finite} set of $M$ bases, if correctly chosen.
A functional map matrix $\C_{kl} \in \R^{M \times M}$ can be then defined \rev{as a replacement for} $T_{kl}$, and \rev{satisfies $T_{kl}(\phi_{l,m_l}) = \sum_{m_k=1}^M (\C_{kl})_{m_k,m_l} \phi_{k,m_k}$}, where $(\C_{kl})_{m_k,m_l}$ denotes the element at the $m_k$-th row and the $m_l$-th column.
We can then re-write \rev{the relation $f_k = T_{kl} (f_l)$} in linear algebra as $\h_k = \C_{kl} \h_l$, where $\h_k := [h_{k,1},...,h_{k,M}]^\top$, $\h_l := [h_{l,1},...,h_{l,M}]^\top$. 
This re-formulation admits various computation tools available for optimization.
We refer readers to a full tutorial provided in \cite{ovsjanikov2016computing} for more in-depth discussions.

\parahead{Discretization on Point Clouds}
As points are samples from the surfaces, we can define real-valued functions on each point cloud (\eg, $\X_k$) as a column vector in $\R^{N_k}$. 
Horizontally stacking the set of $M$ \emph{basis} functions $\{\phi_{k,m}\}_{m=1}^M$ gives a full-rank compact basis matrix $\basis_k \in \R^{N_k \times M}$. 
The pairwise linear functional map matrix $\C_{kl}$, when \rev{left-multiplied with} $\basis_k$, linearly re-combines the bases from $\X_k$ and yields the transferred set of bases $\basis_l$ from $\X_l$ to $\X_k$, \ie, $\basis_k \C_{kl} \approx \Perm_{kl} \basis_l$, where $\Perm_{kl}$ is the point-wise permutation matrix between the two point clouds.
For other \rev{arbitrary functions $\mathbf{f} \in \R^{N_k}$, their coordinates under the bases are given by $\basis_k^+ \mathbf{f}$}, where $\cdot^+$ is the Moore-Penrose pseudo-inverse operator.

\subsection{Generating Putative Correspondences}
\label{subsec:corr}
We estimate deep features $\desc_k \in \R^{N_k \times F}$ for each point cloud in $\Vertex$ using a sparse-convolution-based~\cite{choy20194d} feature descriptor network $\descnet: \X_k \rightarrow \desc_k$.
Given a pair of descriptors $\desc_k$ and $\desc_l$, we construct \rev{the soft permutation matrix $\Perm_{kl}^\de := \mathrm{softmax} (\hat{\Perm}_{kl}^\de) \in \R^{N_k \times N_l}$} and compute the scene flow as follows:
\begin{equation}
\begin{aligned}
    \flow_{kl}^{\de} & := \Perm_{kl}^\de \X_l - \X_k, \\ 
    \rev{(\hat{\Perm}_{kl}^\de)_{ij}} & := \rev{\rev{-}\frac{1}{t^\de} \lVert (\desc_k)_{i:} - (\desc_l)_{j:} \rVert},
\end{aligned}
\end{equation}
where $t^\de$ is a trainable parameter with the initial value of 1.0 and a minimum value of 0.02.
$\mathrm{softmax}(\cdot)$ performs \rev{softmax normalizations over all the rows of $\hat{\Perm}_{kl}^\de$ so that the output $\Perm_{kl}^\de$ becomes row-stochastic (\ie, each row sums to 1)}.
The flow loss $\loss_\mathrm{f}$ introduced in \cref{subsec:loss} will be used to supervise $\descnet$ until convergence.
Remarkably, such a simple strategy, similar to \cite{puy20flot}, already produces the flow $\flow_{kl}^\de$. 
However, such an estimation is corrupted with noise, occlusions, and inconsistencies hence performing a lot worse than our full pipeline.
\rev{This is verified through the `Ours ($\descnet$ only)' baseline in the experiments, where we simply map each point to its nearest neighbor in the space of $\desc_k$.}

After $\descnet$ is trained, it is fixed and we define the correspondence generator as $\corr_\pd: \left( (\X_k, \basis_k ), (\X_l, \basis_l ) \right) \rightarrow ( \basis_k^{(kl)}, \basis_l^{(kl)} )$, which selects \emph{corresponding} rows (\ie, matched points) from the input bases $\basis_k$ and $\basis_l$ (detailed below), and produces $\basis_{k}^{(kl)} \in \R^{I_{kl}\times M}$ and $\basis_{l}^{(kl)} \in \R^{I_{kl}\times M}$, where $I_{kl}$ is the number of matches.
The row indices are determined by a nearest neighbor search performed on the descriptors $\desc_k$ and $\desc_l$ with cross-check, and the rows are matched if the L2 distance of the descriptors is smaller than a \emph{conservative} threshold of 0.3.

\subsection{Computing Basis Functions}
\label{subsec:basis}

The basis functions $\{\basis_k\}$ play a critical role on the power of learned representations. For triangulated shapes where connectivity information is provided, bases can be easily formed via the standard method of decomposing the graph Laplacian~\cite{zhang2010spectral}.
Similarly, for point clouds one can also build a mesh structure such as a k-NN graph or intrinsic Delaunay triangulation~\cite{Sharp:2020:LNT} before computing the bases. 
However, the latter can be problematic because the constructed graph would have no notion of semantics or geodesics, leading to redundant or erroneous function approximation. 
Instead, we propose to learn the basis functions directly from point clouds using a sparse-convolution-based~\cite{choy20194d} neural network $\basisnet: \X_k \rightarrow \basis_k$. Basis functions learned in such manner are both
(1) accurate, \ie focus only on the flow properties we are interested in, and
(2) compact with no redundancies, as will be demonstrated in \cref{sec:exp}, which allows achieving higher accuracy even with a small number of bases.

In order to train $\basisnet$, we compute the optimal functional maps as
\begin{equation}
\label{eq:amin}
\C_{kl}^0 := \argmin_{\C} E_{kl}(\C),
\end{equation}
where
\begin{equation}
\label{eq:ekl}
\begin{aligned}
E_{kl} (\C) & := \sum_{i=1}^{I_{kl}} \rho \left( \lVert \basis_{l,i:}^{(kl)} - \basis_{k,i:}^{(kl)} \C \rVert \right), \\
( \basis_k^{(kl)}, \basis_l^{(kl)} ) & = \corr \left( (\X_k, \basis_k ), (\X_l, \basis_l ) \right).
\end{aligned}
\end{equation}
Here, $\corr(\cdot)$ is a putative correspondence generator detailed earlier. 
The rows of $\basis_{k}^{(kl)}$ and $\basis_{l}^{(kl)}$ are indexed via $(\cdot)_{i:}$.
During training we use the strategy similar to~\cite{bengio2015scheduled} by randomly instantiating $\corr$ either with the ground-truth $\corr_\gt$ (if provided) or the predicted one $\corr_\pd$ with a probability of 50\%. We empirically observe the positive impact of this on final performance. 
\rev{During test time, when ground-truth information is not available, we simply let $\corr = \corr_\pd$.}
$\rho(\cdot)$ is the Huber robust function~\cite{huber1992robust} used for outlier rejection that individually treats each summand $i$.
The scale of $\rho(\cdot)$ is chosen to be 0.05 and empirically we do not observe any significant difference using more advanced adaptive scales such as the median absolute deviations~\cite{hampel1974influence}.
After obtaining $\C_{kl}^0$ we feed it into the method in \cref{subsec:sfffm} to compute the flow vectors, and supervise the networks using the loss function defined later in \cref{subsec:loss}.

\begin{figure}[!t]
\centering
\includegraphics[width=\linewidth]{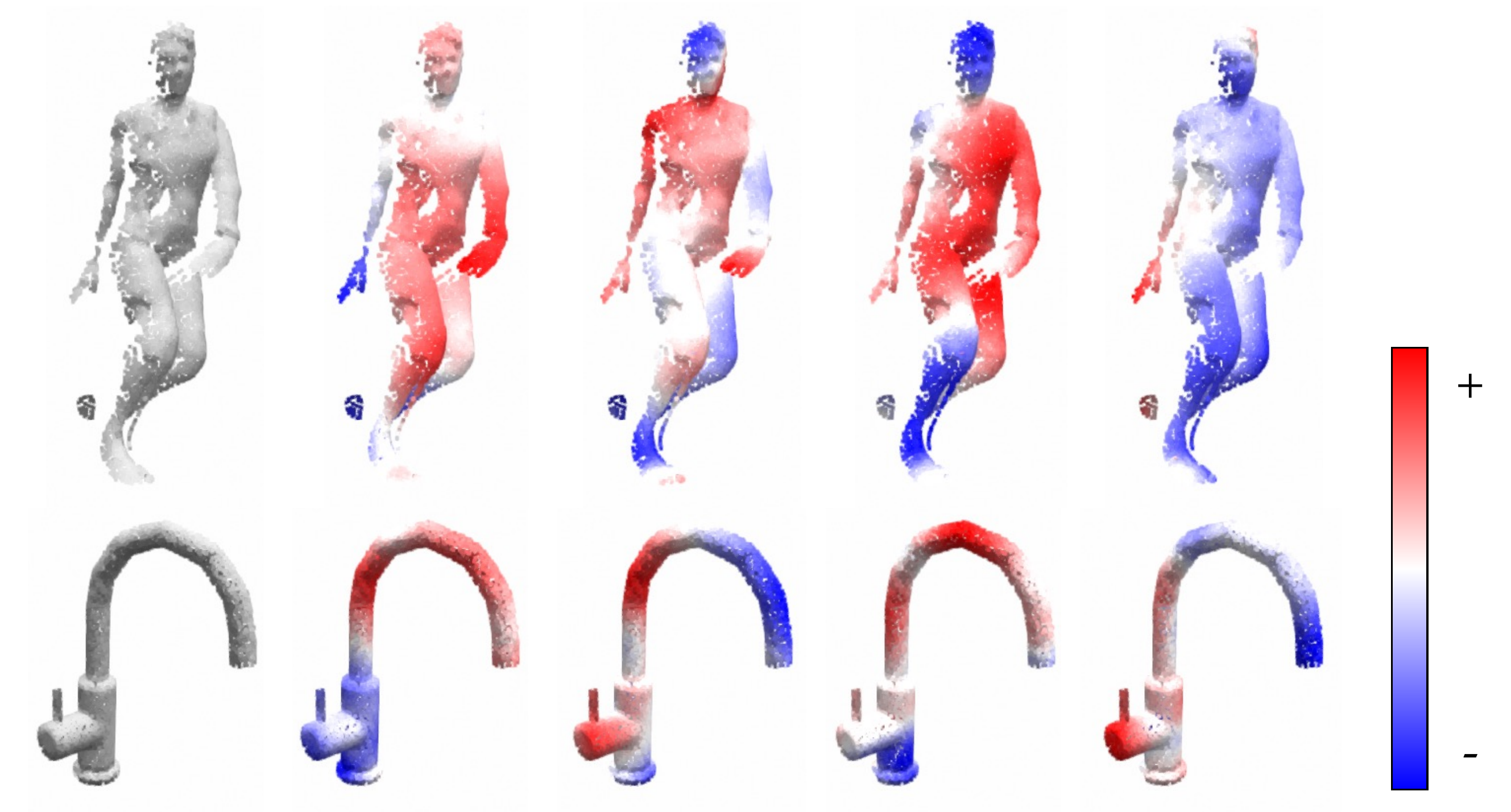}
\vspace{-1.5em}
\caption{\textbf{Learned bases visualization.} We visualize for two shapes (left) a chosen subset of its $M$ bases. The basis range is normalized.}
\vspace{-1em}
\label{fig:basis}
\end{figure}

\subsection{Recovering Flows from Functional Maps}
\label{subsec:sfffm}

Given a computed functional map matrix $\C_{kl}$, which could either be $\C_{kl}^0$ or the optimized one $\C_{kl}^\star$ from \cref{sec:sync}, one can again compute the soft permutation matrix $\Perm_{kl} \rev{:= \mathrm{softmax} (\hat{\Perm}_{kl})}$ between the two clouds and extract the flow as:
\begin{equation}
\label{eq:flownn}
\begin{aligned}
    \flow_{kl}^{\nn} & := \Perm_{kl} \X_l - \X_k, \\ \rev{(\hat{\Perm}_{kl})_{ij}} & := \rev{\rev{-}\frac{1}{t} \lVert (\basis_k \C_{kl})_{i:} - (\basis_l)_{j:} \rVert}.
\end{aligned}
\end{equation}
However, as pointed out before, $\Perm_{kl}$ is row-stochastic and the warped position $\Perm_{kl} \X_l$ hence never extends beyond the \emph{convex hull} of $\X_l$, which is detrimental for computing flows for occluded points.
An alternative way to compute the flow is to treat the target positions $\X_l$ \emph{themselves} as three functions and map their coordinates via $\C_{kl}$, evaluated as:
\begin{equation}
\label{eq:flowbs}
    \flow_{kl}^{\bs} := \basis_k \C_{kl} \basis_l^+ \X_l - \X_k.
\end{equation}

Note that the ground-truth mapping already implies $\basis_k \C_{kl} \approx \Perm_{kl} \basis_l$ as noted in \cref{subsec:fmap_basic}. However, the two methods for computing flows are fundamentally different, conceptually:
$\flow_{kl}^{\bs}$ elegantly handles the issue of occlusions because the mapping allows us to map unknown positions to valid ranges given the continuity of the bases, leading to a learned extrapolation scheme via band-limited regularization.
Nevertheless, the truncated basis may over-regularize the positions in geometrically-complicated parts and limit the representation power of the flows. In contrast, $\flow_{kl}^{\nn}$ can perform well on those regions thanks to notion of point-wise correspondences.

Hence, we combine the best of the two worlds by an additional refinement network $\refnet$ which digests the concatenation of the two, along with $\X_k$, and outputs the residual:
\begin{equation}
\label{eq:flowfin}
    \flow_{kl} := \flow_{kl}^{\nn} + \refnet ([\X_k, \flow_{kl}^{\bs}, \flow_{kl}^{\nn}]).
\end{equation}

The refinement network $\refnet$ is instantiated with a sparse-convolution-based network, which works in a coarse-to-fine fashion by first aligning the general cloud structure with $\flow^\bs_{kl}$ and then fixing small-scale detailed errors by considering the information from $\flow^\nn_{kl}$.
The smoothness property of such networks also helps prune predicted outliers as already demonstrated by previous works (\eg \cite{puy20flot}).
A comparison between the different flows is shown in \cref{fig:occlusion}.

\subsection{Loss and Training}
\label{subsec:loss}
We train our network in two steps, where:
(1) We train the correspondence generator $\corr$ described in \cref{subsec:corr} using a \emph{flow loss} $\loss_\mathrm{f}$ only, and
(2) we jointly train $\basisnet$ and $\refnet$ (\cref{subsec:basis} and \cref{subsec:sfffm}) end-to-end adding the \emph{consistency loss} $\loss_\mathrm{c}$, \ie: 
\begin{equation}
    \loss_\mathrm{f} + \lambda_\mathrm{c} \loss_\mathrm{c}.
\end{equation}

Notably, the $\mathrm{argmin}$ operator in \cref{eq:amin} is differentiated by computing the solution with Iteratively Reweighted Least Squares (IRLS) and unrolling the re-weighting iterations.
We provide the details of this algorithm in Appendix~\refap{C}.

\parahead{Flow Loss $\loss_\mathrm{f}$}
If ground-truth annotations $\flow_{kl}^\gt$ are provided, we can directly use:
\begin{equation}
    \loss_\mathrm{f,sup} := \lVert \flow_{kl} - \flow_{kl}^{\mathrm{gt}} \rVert_\fro^2,
\end{equation}
where $\lVert \cdot\rVert_\fro$ denotes Frobenius norm.
Otherwise, we use the self-supervised loss inspired from~\cite{wu2019pointpwc}:
\begin{equation}
    \loss_\mathrm{f,unsup} := \sum_{\mathrm{type} \in \{ \mathrm{chamfer,smooth,lap} \}} \lambda_\mathrm{type} \rev{\loss_\mathrm{type}},
\end{equation}
where for conciseness, the respective loss terms $\loss_\mathrm{type}$, the weights being $\lambda_\mathrm{type}$, are defined in Appendix~\refap{A}.
We evaluate both settings using either $\loss_\mathrm{f,sup}$ or $\loss_\mathrm{f,unsup}$ in our experiments (see \cref{subsec:dataset} for details). 
The more general $\loss_\mathrm{f}$ refers to both for convenience.

\parahead{Consistency Loss $\loss_\mathrm{c}$}
We impose the pairwise consistency loss~\cite{roufosse2019unsupervised} as:
\begin{equation}
    \loss_\mathrm{c} := \lVert \C_{kl}^0 \C_{lk}^0 - \Id_M \rVert_\fro^2,
\end{equation}
where $\Id_M$ is the identity matrix with size $M \times M$.
This loss regularizes the compound mapping from functions on $\X_k$ to functions on $\X_l$ and its backward forms an identity, ensuring a local consistency of the learned bases.
Additionally, we tried other regularization terms over the bases, such as Laplacian commutativity~\cite{ovsjanikov2012functional}\footnote{This means to project the Laplacian operator $\mathbf{L}$ to the current bases via $\basis^+\mathbf{L}\basis$.} or descriptor preservation~\cite{Nogneng2017InformativeDP} but empirically no improvement is observed.


\subsection{Discussions}
\label{subsec:discuss}

\parahead{Further Notice of $\basis$ and $\C$}
Examples of the learned basis functions are visualized in \cref{fig:basis}.
Remarkably, the learned bases $\basis$ and the functional map matrix $\C$ are not constrained to a particular geometric structure, \eg, orthonormality $\basis^\top \basis = \Id$ or orthogonality $\C \C^\top = \Id$, which hold only under isometric deformations with full geometry for meshes.
The coarse-to-fine structure of $\basis$ is not enforced either: the network finds its own way of representing high-frequency bases via learning small differences.
Empirically we find that enforcing any of these regularizations harms the performance.
On the other hand, unlike most existing works (\eg, \cite{donati2020deep,marin2020correspondence}) that formulates \cref{eq:ekl} by projecting \textit{probe functions} $\mathbf{P}_k,\mathbf{P}_l$ via $\C_{kl}^0 := \argmin_{\C} \lVert \C \basis_l^+ \mathbf{P}_l - \basis_k^+ \mathbf{P}_k \rVert $, our method is more robust due to the per-match outlier filtering and gets rid of the trade-off between the flexibility of probe functions and the compactness of the bases for flow estimation. 

\parahead{Special Case: Affinity Bases}
For large-scale scenes with multiple moving \emph{rigid} objects as often observed in the case of autonomous driving, the resulting flow can be effectively regularized by constraining the basis structure.
Specifically, if the segmentation of $S$ rigid bodies (each indexed by $s$) is known beforehand (\eg, acquired via \cite{Gojcic2021WeaklySL} or semantic segmentations), instead of learned ones, we can manually define the basis functions as:
\begin{equation}
    \forall s, (\basis_{k})_{:,4s:4s+4} = \diag(\mathbf{G}_k = s) \begin{bmatrix}
        \X_k & \one
    \end{bmatrix},
\end{equation}
\vspace{-1em}
\setlength{\columnsep}{5pt}
\setlength{\intextsep}{0pt}
\begin{wrapfigure}[6]{r}{0.33\linewidth}
\vspace{-1ex}
\begin{center}
\includegraphics[width=\linewidth]{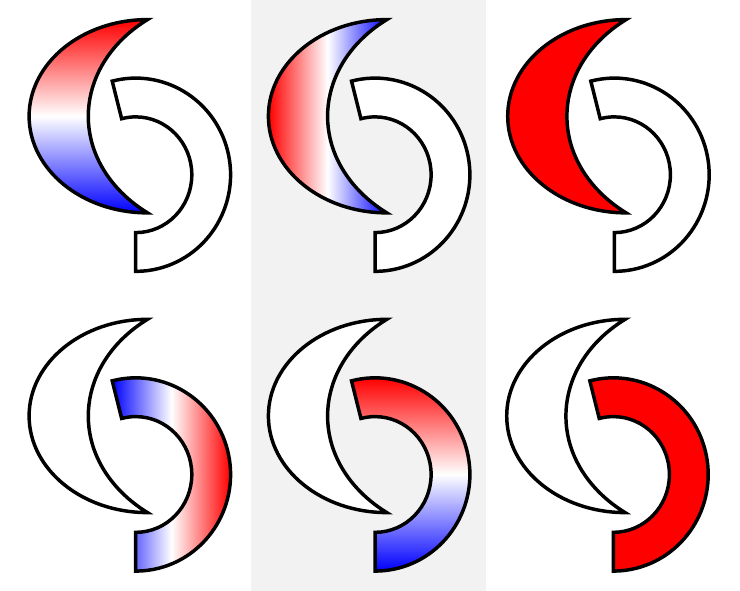}
\end{center}
\end{wrapfigure}
where $\diag(\cdot)$ is the diagonal matrix of an indicator vector, $\mathbf{G}_k \in \R^{N_k}$ are the rigid body indices corresponding to $\X_k$ and the final bases $\basis_k \in \R^{N_k \times 4S}$ are split into $S$ blocks (as visualized in the inset as a 2D example).
Note that the resulting map $\C_{kl} \in \R^{4S \times 4S}$ not only shows the affine transformation matrices in its $4\times 4$ sub-blocks, but the sparsity of the sub-blocks also reveals rigid-body-level permutations.
As a demonstration of the effectiveness of such bases, we provide evaluations on driving scenes in \cref{subsec:exp:kitti}.

\section{Functional Synchronization}
\label{sec:sync}

For a geometrically coherent outcome, we further ask the loops in $\Graph$ to be closed. 
\rev{Note that this step happens only during test time after the pairwise networks are trained.}
In our setting of functional mappings, this amounts to enforcing the cycle-consistency~\cite{birdal2021quantum}:
\begin{equation}
    \C_{k_1 k_2} \dots \C_{k_{p-1} k_p}\C_{k_p k_1} = \Id, \quad \forall (k_1, \dots k_p)\in \mathcal{C}(\Graph).
\end{equation}
It can be easily shown by construction that the above formulation is equivalent to defining a set of $V$ \emph{canonical} functions $\Hfunc_k = [ \h_k^1, \dots, \h_k^V ] \in \R^{M \times V} $ on each point cloud $\X_k$, expressed as the coordinates under $\basis_k$. 
By enforcing the canonical functions to evaluate the same for the corresponding regions on different shapes, we define the following cycle consistency energy:
\begin{equation}
    E_{\mathrm{cycle}}(\{\Hfunc_k\}, \{\C_{kl}\}) := \sum_{(k,l)\in\Edge} \lVert \Hfunc_k - \C_{kl} \Hfunc_l \rVert_F^2.
\end{equation}

Note that a naïve solution will collapse to $\Hfunc_k = \zero\,\,\forall k$, so an additional orthonormality constraint is added as $\Hfunc^\top \Hfunc = \Id_V$, where $\Hfunc := [ \Hfunc_1^\top, \dots, \Hfunc_K^\top ]^\top \in \R^{KM \times V}$. 
This constraint eliminates the trivial solution by \emph{fixing the gauge} and encourages a larger functional space spanned by the canonical functions.

Additionally, we define the data term as the summation of all pairwise energies:
\begin{equation}
\label{eq:data}
    E_{\mathrm{data}}(\{\C_{kl}\}) := \sum_{(k,l)\in\Edge} E_{kl} (\C_{kl}),
\end{equation}
with $E_{kl}$ defined in \cref{eq:ekl}.
\rev{Recall that during test time we set $\corr$ (needed in $E_{kl}$) to $\corr_\pd$ without any reliance on ground-truth information.}
The solution to the problem 
\begin{equation}
\label{eq:sync}
    \argmin_{\{\Hfunc_k\}, \{\C_{kl}\}} E_{\mathrm{cycle}}(\{\Hfunc_k\}, \{\C_{kl}\}) + E_{\mathrm{data}}(\{\C_{kl}\})
\end{equation}
gives us the synchronized functional maps $\{\C_{kl}^\star\}$ as well as one possible set of canonical functions. 
After the optimization, we re-use the technique described in \cref{subsec:sfffm} to recover $\Fset$ as our final output, by feeding in the optimized result $\{\C_{kl}^\star\}$ instead of $\{\C_{kl}^0\}$.

\parahead{Analysis}
In our formulation, each summand $E_{kl}$ in $E_\mathrm{data}$ of \cref{eq:data} now contains the robust function $\rho(\cdot)$ whose residual not only reflects the alignment with the current pairwise map, but also the consistency among all the maps in $\Graph$, related by $\{\Hfunc_k\}$.
Remarkably, compared to existing point-based synchronization techniques (\eg \cite{huang2021multibodysync}), the use of functional maps is \emph{lightweight} reducing the number of variables from the order of $O(N^2)$ to $O(M^2)$ where $M \ll N$. Besides. its additional degree-of-freedoms endow us with more \emph{flexibility} in representing non-rigid transformations compared to, \eg~\cite{gojcic2020learning}.

\parahead{Numerical Optimization}
\rev{Inspired by \cite{wang2013image}, we use an alternating method to solve~\cref{eq:sync}. Specifically, we iterate over optimizing $\{\Hfunc_k\}$ with fixed $\{\C_{kl}\}$ and optimizing $\{\C_{kl}\}$ with fixed $\{\Hfunc_k\}$ until convergence. Details of the solver are deferred to Appendix~\refap{B}.}

\parahead{Basis Preconditioning}
While we do not regularize the columns of $\basis_k$ to be orthogonal during the training of $\basisnet$, optimizations related to it can be highly ill-posed due to the emergence of near-parallel columns.
This is due to the use of pseudo-inverse that implicitly factors out the high-frequency bases from the network output, which by nature tends to produce over-smoothed results. 
Such a phenomenon, though does not affect network training without synchronization (\ie \cref{sec:pw}), jeopardizes the convergence and stability when jointly optimizing~\cref{eq:sync}.
In order to mitigate this issue, during test time, we explicitly leverage orthogonality as a means of preconditioning by performing the singular value decomposition (SVD) as $\basis_k = \mathbf{U} \mathbf{\Sigma} \mathbf{V}^\top$ and replace $\basis_k$ with $\mathbf{U}$ as the new set of basis functions.
Note that \cref{eq:flownn} and \cref{eq:flowbs} are calibrated during training to engulf the original basis scale from the raw network output, hence the multiplier $\mathbf{\Sigma} \mathbf{V}^\top$ should be applied back after synchronization to recover the final 3D flow vectors.

%% file: sections/grp4-exp.tex
\section{Experiments}
\label{sec:exp}

\subsection{Dataset and Settings}
\label{subsec:dataset}
\vspace{-0.5em}

\parahead{Datasets}
State-of-the-art methods for non-rigid registration or matching usually evaluate on different datasets with different settings. 
In order to provide a comprehensive and fair evaluation, we experiment with a full spectrum of possible input data, ranging from partial to full point clouds, from non-rigid human/clothes/animals to multi-body articulated objects, and from synthetic to real-world scenes. 
To this end, we extend the following datasets~\cite{cape1,cape2,Li20214DCompleteNM,Bozic2020DeepDeformLN, Xiang_2020_SAPIEN} to the task of multi-way non-rigid registration (hence the prefix \textbf{M}ulti-\textbf{P}oint \textbf{C}loud used below).
\rev{For all the datasets we keep only the points from the foreground deforming objects.}
\begin{itemize}[topsep=0pt,leftmargin=*]
\setlength{\itemsep}{0pt}
\setlength{\parskip}{0pt}
    \item \textbf{\dcape} is sampled from CAPE~\cite{cape1,cape2} dataset, which contains scanned sequences of clothed human using a commercial 3dMD scanner. \rev{Raw scans are cropped using a fixed-height horizontal plane to remove the static ground points}. We randomly sample non-consecutive $K$-frame snippets from the sequences and obtain the ground-truth flow from the fitted template mesh skinned via a Linear Blend Skinning~\cite{skinningcourse:2014} scheme.
    \item \textbf{\ddtfd} is from the DeformingThings4D \cite{Li20214DCompleteNM} dataset, with hundreds of dynamic sequences of humanoids and animals whose skeletons and motions are designed by experts. We extract the partial point clouds by synthetically scanning the object under different motions. \rev{We ensure that the \emph{(instance, action)} tuples never overlap when we create the splits.}
    \item \textbf{\ddd}~\cite{Bozic2020DeepDeformLN} contains partial views captured with a real RGB-D camera. The scene flows are provided by the original dataset via non-rigidly tracking the temporally densely-sampled frames. \rev{The provided object masks are used to remove the static background points.} \rev{We use the original validation set as our test set and split the original training set for training and validation, without instance overlap between the splits}.
    \item \textbf{\dsapien}~\cite{Xiang_2020_SAPIEN} consists of simulated, realistic, articulated object-level models that are commonly observed in daily life. We follow the data generation and sampling strategies implemented in \cite{huang2021multibodysync}, \rev{which ensure that there is no instance overlap between the splits}.
\end{itemize}
Statistics of the dataset splits are shown in~\cref{tbl:dataset}.

\parahead{Metrics}
The main metric for evaluating our flow quality is the (1) \epe, also termed as Mean Absolute Error (MAE), or End-Point Error (EPE), \ie, the average norm of the flow error vectors over all points.
We also include metrics from the scene flow~\cite{liu2019flownet3d} community but adapt their thresholds to our object-level setting.
Specifically, we add:
(2) 3D Accuracy Strict (AccS), the percentage of points whose relative error $<5\%$ or $<2$cm,
(3) 3D Accuracy Relaxed (AccR), the percentage of points whose relative error $<10\%$ or $<5$cm, and
(4) Outlier Ratio, the percentage of points whose relative error $>30\%$.

\input{tables/dataset}


In datasets with occlusions, we additionally compute all the metrics both for non-occluded points and the full point clouds.
We subsequently collect the statistics among all the $K(K-1)$ pairs of the metrics and report the mean value and standard deviation ($\pm$), the latter of which shows the consistency of the estimates among all individual pairs, as desired in the multiway setting.

\input{tables/cape-full}

\begin{figure*}[!t]
\centering
\vspace{-0.5em}
\includegraphics[width=0.95\linewidth]{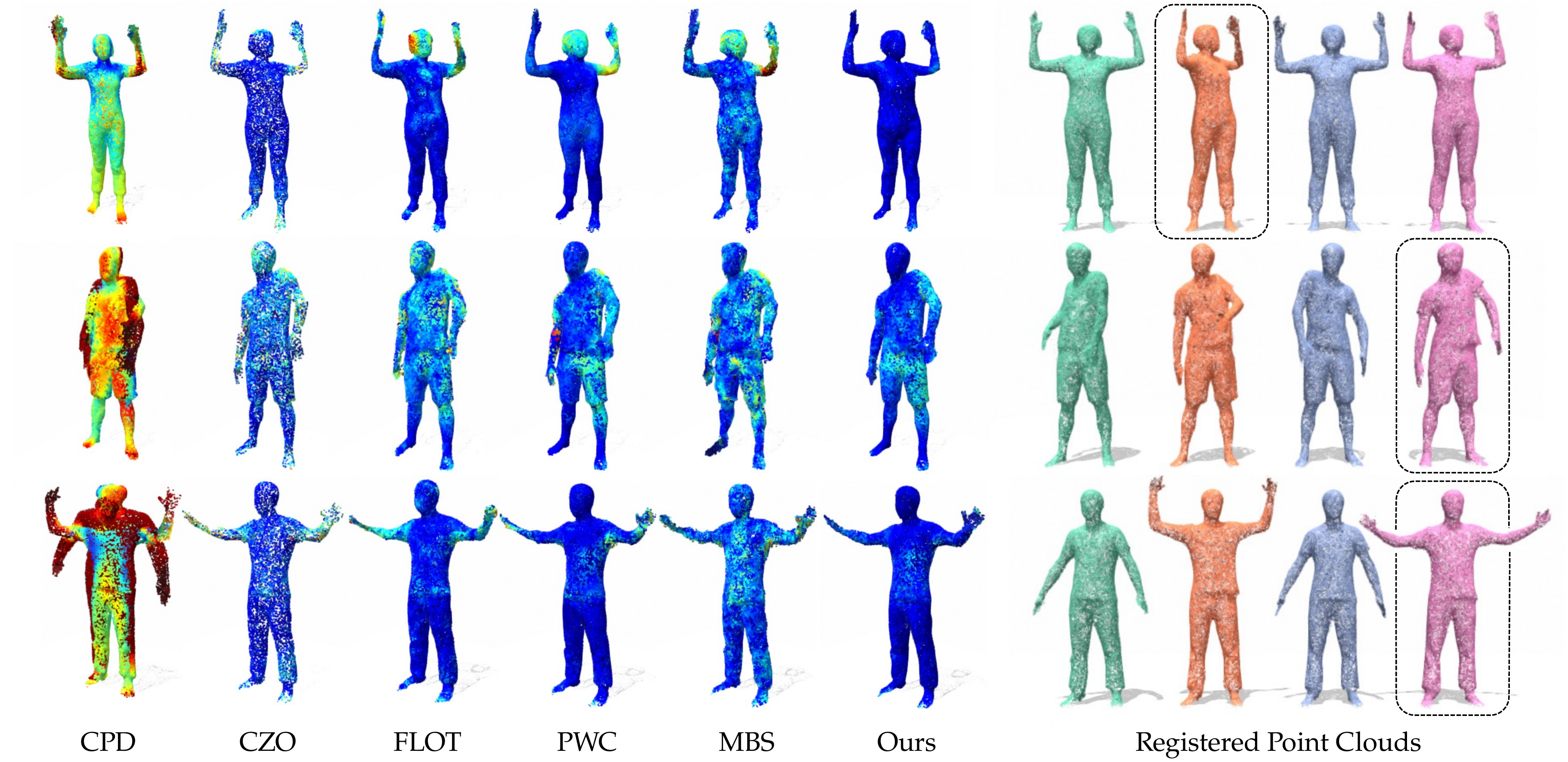}
\vspace{-1em}
\caption{\textbf{Comparison of fully-supervised \dcape.} Left: Color-coded per-point errors (0cm~\jetbar~10cm) are shown on the registered point clouds by warping all points to a selected pose. Right: Registered point clouds using our inferred 3D flows. Dashed rectangles are the views to which we warp other point clouds on the left side.}
\label{fig:cape}
\vspace{-1em}
\end{figure*}

\parahead{Baselines}
We compare the performance of \name~against a number of baselines grouped by the learning strategy.
Hereinafter, we use the underlined names for brevity:
\begin{itemize}[topsep=0pt,leftmargin=*]
\setlength{\itemsep}{0pt}
\setlength{\parskip}{0pt}
\item \textbf{Non-learned}: Coherent Point Drift method (\underline{CPD})~\cite{myronenko2010cpd}, 
Group-wise unnormalized information potential with Rényi's entropy (\underline{GUIP})\cite{Giraldo2017GroupWisePR}, 
and Non-Parametric Part (\underline{NPP})~\cite{hayden2020nonparametric} using up to 25 RJ-MCMC sample steps; 
\item \textbf{Supervised}: Consistent ZoomOut (\underline{CZO})~\cite{huang2020consistent} where we replace the original Laplacian-Beltrami basis functions with the non-manifold one in~\cite{Sharp:2020:LNT} (the number of bases is set to 80), 
scene flow estimation methods including \underline{FLOT}~\cite{puy20flot} and PointPWC-Net (\underline{PWC})~\cite{wu2019pointpwc}, MultiBodySync (\underline{MBS})~\cite{huang2021multibodysync} where for non-rigid case we remove the motion segmentation module and apply only the weighted permutation synchronization with downsampled 512 points, \rev{and \underline{EmbAlign}~\cite{marin2020correspondence} which also estimates point cloud bases.}
\item \textbf{Self-supervised}: 
PointPWC-Net (\underline{PWC-U})~\cite{wu2019pointpwc} with the unsupervised loss, and RMA-Net~(\underline{RMA})\cite{feng2021recurrent} with the proposed differentiable rendering loss.
\end{itemize}

\rev{For \dcape{} where full human bodies are captured, we additionally compare to template-based methods \underline{3D-CODED}~\cite{groueix20183d} and \underline{PTF}~\cite{wang2021locally}, as well as deep-functional-map-based methods \underline{GeomFMaps}~\cite{donati2020deep} (supervised) and \underline{DeepShells}~\cite{eisenberger2020deep} (self-supervised), where bases are replaced with \cite{Sharp:2020:LNT}.}
For baselines that do not have a principled way of  handling multiple inputs, we treat each pair of the input point clouds independently. 
Readers are referred to the appendix for more descriptions of the baseline settings.

\parahead{Parameter Settings}
The networks $\descnet$, $\basisnet$, and $\refnet$ are all trained using AdamW~\cite{loshchilov2017decoupled} optimizer with the initial learning rate of $10^{-3}$ and a 30$\%$ decay every 50k samples.
$\descnet$ and $\basisnet$ are both 4-layer U-Net~\cite{choy2020deep} with 32, 96, 64, 192 channels and 32, 64, 128, 256 channels respectively, while $\refnet$ is a 2-layer small network with 32 and 64 channels.
All networks use InstanceNorm~\cite{ulyanov2016instance} as normalization.
\rev{Further details about the network architectures are illustrated in Appendix~\refap{D}.}
The initial $t$ in \cref{eq:flownn} is chosen to be 0.1, and the weight for $\mathcal{L}_\mathrm{c}$ is $\lambda_\mathrm{c} =5 \times 10^{-3}$.
For our self-supervised training scheme, we balance the Chamfer/Laplacian/smoothness losses with the weights 1.0/5.0/1.0, respectively.
For benchmarking we choose $K=4$ and use $M=24$ for all of our experiments except for the one in \cref{subsec:exp:kitti}.
$V$ is set to be $M-4$ for \dcape~and $M-2$ for all the other datasets.

\input{tables/faust}

\subsection{Evaluations on \dcape}
\label{subsec:exp:cape}

\dcape{} is mainly used to measure the fitting tightness and the deformation quality of the warped point clouds.
As shown in \cref{tbl:cape-full}, our method achieves a significant performance boost, reaching a $21.1\%$ and $16.2\%$ lower error compared to the nearest baseline under full supervision and self-supervised scenarios, respectively.

\begin{figure*}[!t]
\centering
\includegraphics[width=0.98\linewidth]{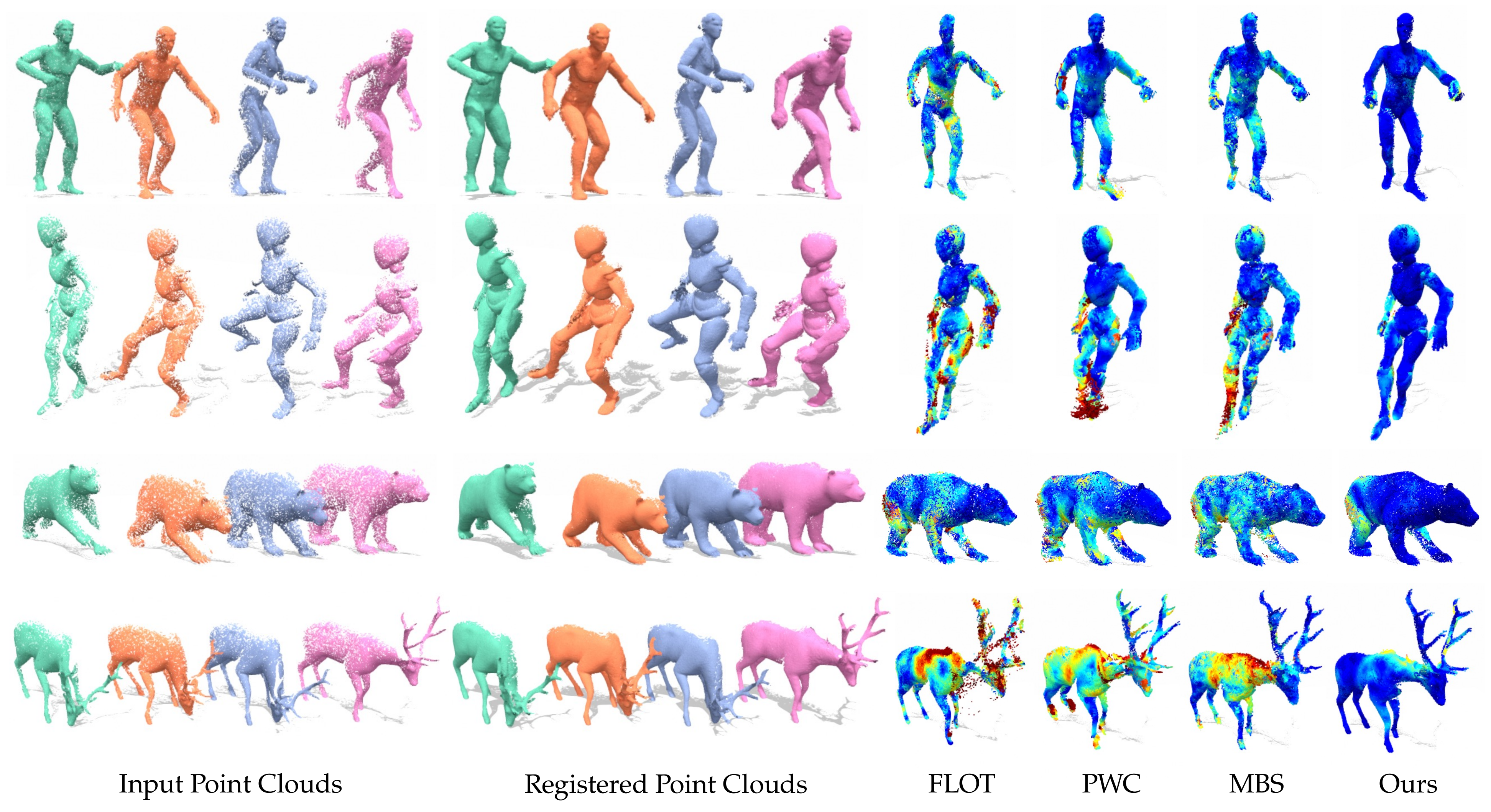}
\vspace{-1em}
\caption{\textbf{Comparison of different methods on the \ddtfd~dataset.} The rightmost four columns show color-coded per-point errors (0cm~\jetbar~10cm) on the registered point clouds by warping all points to a selected pose.}
\vspace{-1em}
\label{fig:dt4d-combined}
\end{figure*}

\input{tables/dt4d-full}

Referring to the visualizations in \cref{fig:cape}, \name~is most effective in point clouds with large changes, where traditional methods fail either due to wrong associations caused by matching ambiguity or insufficient degrees of warps.
For learning-based methods: 
3D-CODED and PTF rely on predefined canonical human shape prior, which prevents them from guaranteeing a tight fitting to the input points.
CZO builds strict point-level correspondences. 
Hence, it is not capable of \emph{densifying} the points and is error-prone once the number of outliers from $\descnet$ is large.
\rev{Moreover, the approximate point cloud bases, used also in GeomFMaps and DeepShells, introduce further drifts.
While PWC and FLOT are trained using dense flow annotations and are free from large systematic errors, they still yield noisy estimates and result in shrunk artifacts in the detailed region.
Comparably, we use the limited-band bases to effectively regularize the motions, while preserving local rigidity and details. 
By explicitly projecting point-level \emph{matches} onto smooth functions, the generated warping is more accurate and robust to noise.
This is in contrast to EmbAlign, where the descriptors are globally projected onto the bases, largely limiting the distinctiveness of the point features.}
In the unsupervised scenario, we are able to outperform PWC-U and RMA thanks to the superiority of our matching representation and the proposed learning strategy.
Moreover, with the help of the synchronization module, we reach a better estimation by pruning inconsistent matches from the graph, reducing the standard deviation of the L2 error by over 5.8\%.

\parahead{FAUST Challenge}
\rev{To reinforce our results on CAPE, we test our pairwise registration module (without synchronization) on the online benchmark FAUST~\cite{bogo2014faust}, where ground-truth information is not accessible. We perform the evaluation on the intra-subject challenge, which measures the registration error between 60 pairs of shapes, where each pair consists of two full meshes of the same subject in different poses.
We directly test our pretrained model without further finetuning on the points sampled from the input meshes. The results are shown in \cref{tbl:faust}. 
Our method ranks 3rd (as of Mar. 2022) among all published methods. 
The two methods that surpass ours either require manually-labelled landmarks \emph{during test time} (DHNN~\cite{jiang2020disentangled}), or rely on artist-designed canonicalization of a predefined human model (DVM~\cite{kim2021deep}). We are free of the above assumptions and output the more general scene flow representation applicable to different subject categories.}



\begin{figure}[tbp]
\centering
\includegraphics[width=\linewidth]{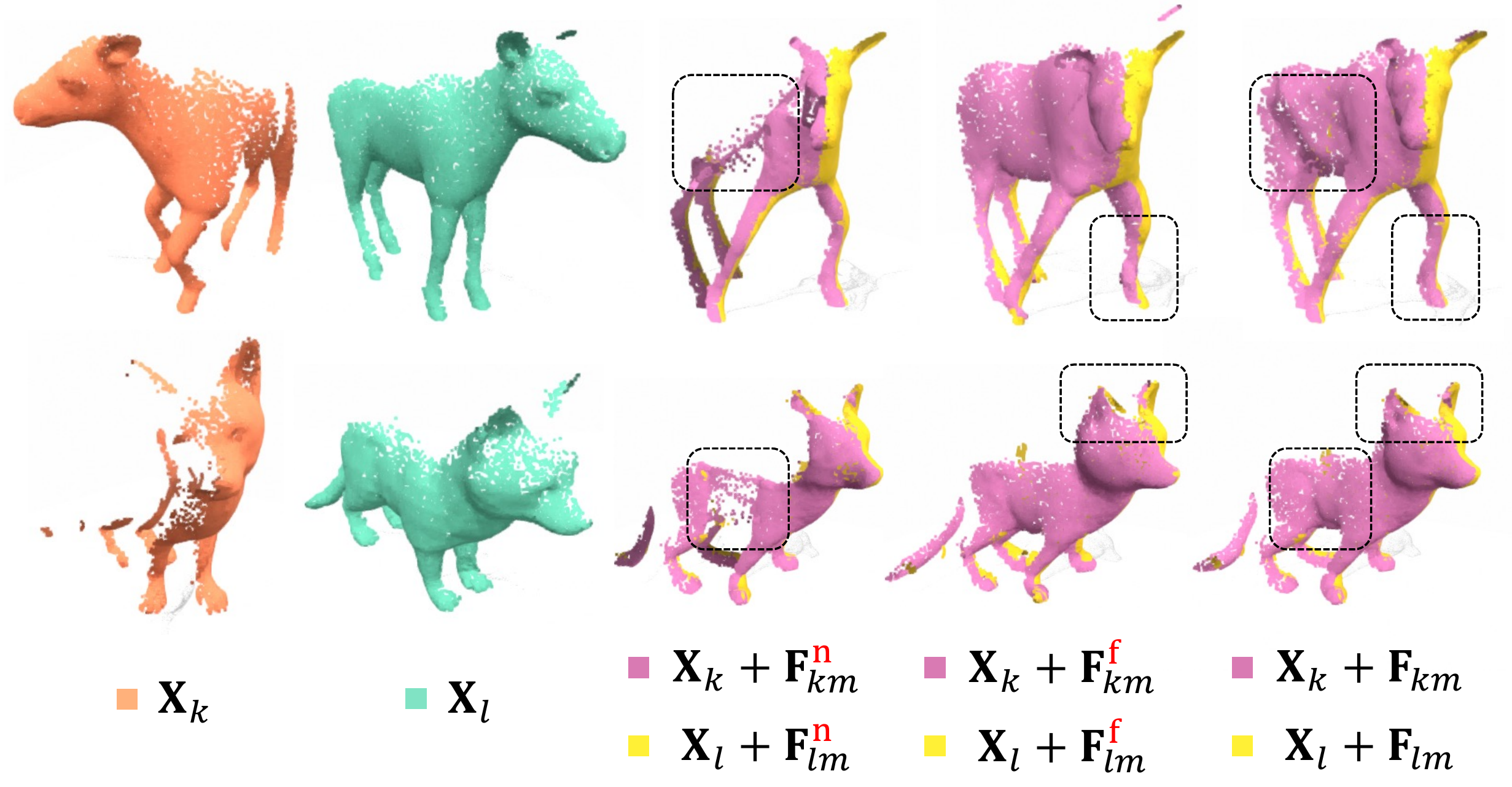}
\vspace{-2em}
\caption{\textbf{Handling low-overlap samples}. We warp two views ($\X_k, \X_l$) to the third view ($\X_m$) using different strategies for flow computation from the functional map matrix (See \cref{subsec:sfffm} for notations and analysis). Dashed rectangles highlight the differences.}
\vspace{-1.5em}
\label{fig:occlusion}
\end{figure}

\subsection{Evaluations on \ddtfd}
\label{subsec:exp:dt4d}

Scanned by a virtual camera, \ddtfd~contains challenging partialities and occlusions, the handling which is an important yet often missing ingredient from many methods.
Input shapes that overlap only partially can cause catastrophic failure for some methods that are incapable of modeling the flow into \emph{empty space}, especially when large deformations are present.
Somewhat surprisingly, \cref{tbl:dt4d-full}, shows that none of the SoTA methods can effectively deal with the hard problem setting presented in this dataset.
On the one hand, methods like CZO and FLOT build the point-level correlation matrix (\eg permutations), which results in large errors on the boundaries of the point cloud.
On the other hand, the motion regularization implicitly introduced in CPD and GUIP cannot be easily adapted to the various kinds of input shapes.
Moreover, the utilization of consistency on the level of points renders MBS inapplicable under such scenarios.

Comparably, as also shown in \cref{fig:dt4d-combined}, we remark that in our method, the learning of the basis functions 
constructs a shared pattern of similar shapes under different views, hence producing a good functional extrapolation of point coordinates in a learned manner.
By not operating on the level of points, our synchronization module bypasses the occlusion problem via aligning the globally-defined basis functions, successfully reducing the error by 5.4\% for the full point cloud.
Further challenging examples that can be successfully tackled by our method are visualized in \cref{fig:occlusion}.


\subsection{Evaluations on \ddd}
\label{subsec:exp:dd}

\begin{figure}[!t]
\centering
\includegraphics[width=\linewidth]{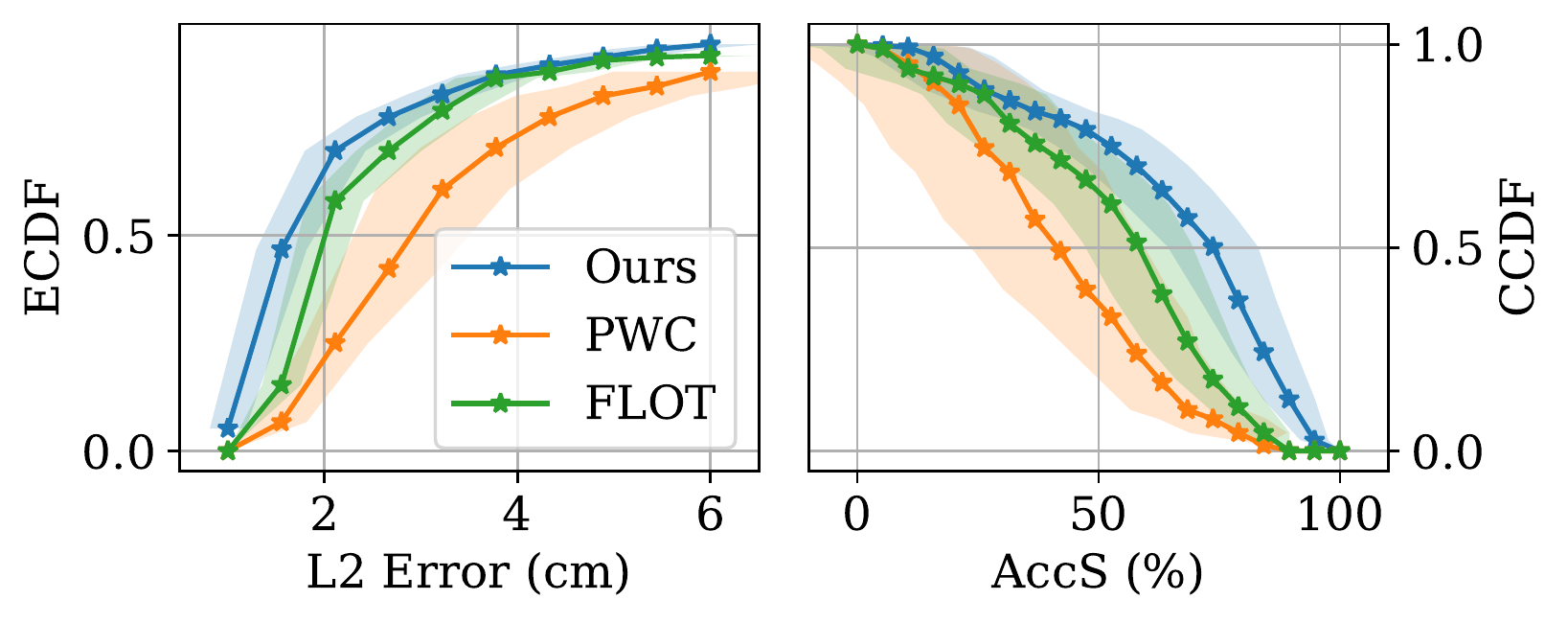}
\vspace{-2em}
\caption{\textbf{Quantitative comparison on \ddd~dataset.} Left and right subfigures show the empirical cumulative distribution function (ECDF) and the complementary cumulative distribution function (CCDF) of \epe~and AccS metrics. Higher curve is better. Shaded areas denote the standard deviations of the respective metrics among views.}
\label{fig:dd-stats}
\vspace{-0.5em}
\end{figure}

As shown in \cref{fig:dd-stats}, our method consistently outperforms scene-flow based methods (19.4\% / 29.2\% lower \epe~and 2.9\% / 11.3\% higher AccS than FLOT / PWC, resp.), even with the presence of non-negligible real-world noise, holes, and large motions.
We show the application of dynamic reconstruction from sparse views (\rev{usually hundreds of frames apart as detailed in the appendix}) in \cref{fig:dd} and found that our learned basis functions are smooth and robust, despite the difficulties in \ddd.

\subsection{Evaluations on \dsapien}

\begin{figure*}[!t]
\centering
\includegraphics[width=0.99\linewidth]{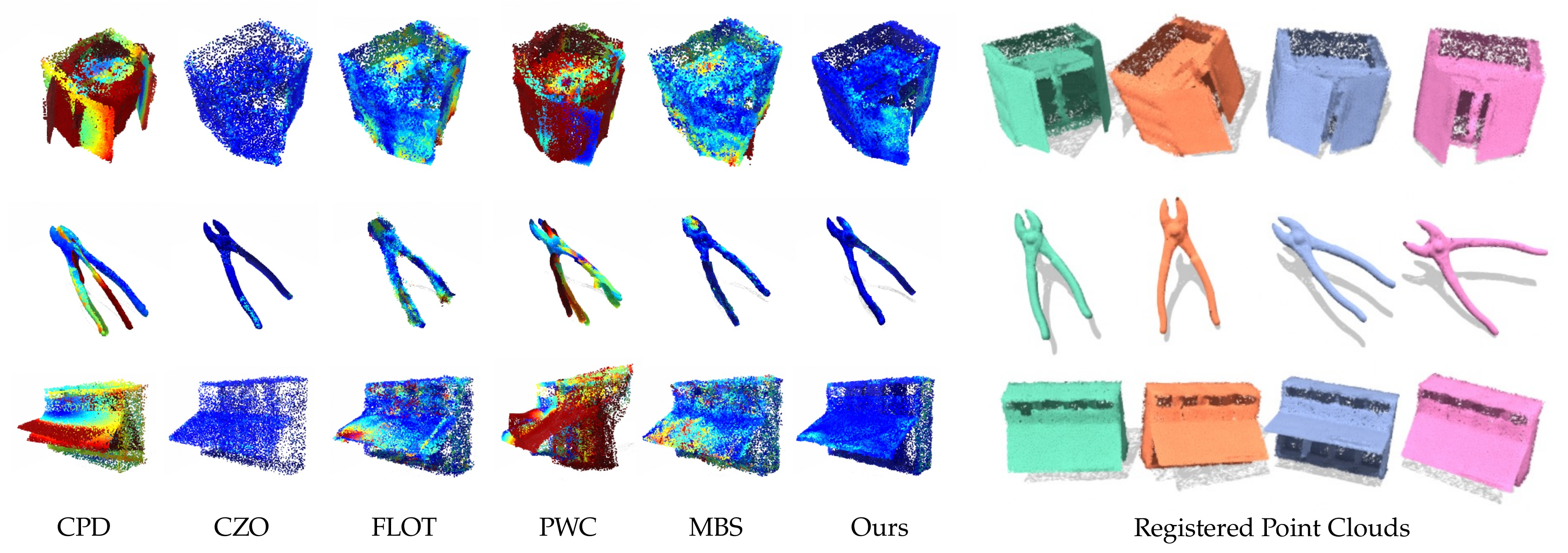}
\vspace{-1em}
\caption{\textbf{Comparison of fully-supervised \dsapien.} Left: Color-coded per-point errors (0.0~\jetbar~0.1) are shown on the registered point clouds by warping all points to a selected pose. Right: Registered point clouds using our inferred scene flows for all input poses.}
\label{fig:sapien}
\vspace{-0.5em}
\end{figure*}

Quantitative and qualitative evaluations on \dsapien{} are shown in \cref{tbl:sapien-full,fig:sapien}.
Note, \dsapien~contains no occlusions and only small \emph{local} deformations. This allows~\cite{Sharp:2020:LNT} to accurately estimate the point cloud bases, making CZO the most effective out of all the competitors.
MBS is specially designed to handle multi-scan multi-body case, and already decreases the PWC backbone error by half thanks to the rigidity prior. However, the permutation and segmentation synchronization modules of MBS limit the maximum number of input points to hundreds coarsening the segmentation and causing a large overall error.
As visualized earlier in \cref{fig:basis}, the basis functions learned from our method already contain motion cues whose patterns are similar to moving part segmentations.
Hence, we conclude that our good performance stems mainly from the explicit usage of bases and the synchronization module, which itself contributes an 11.2\% and 30.3\% improvement over mean \epe~and its variance, providing a better and more consistent flow estimation.

\subsection{Affinity Bases on LidarKITTI}
\label{subsec:exp:kitti}
We now conduct an experiment on the cluttered scenes such as roads in autonomous driving to demonstrate the broad applicability of \name. To this end, we use the LidarKITTI~\cite{Gojcic2021WeaklySL} dataset with ground points removed.
Note that this dataset contains only pairwise Lidar frames, hence the synchronization module is not applied.
\rev{The correspondence estimator $\corr$ is trained on FlyingThings3D~\cite{mayer2016large} dataset using the method described in \cref{subsec:corr}} and the bases are computed according to the \emph{affinity bases} strategy in \cref{subsec:discuss} with the foreground segmentation module from \cite{Gojcic2021WeaklySL}.
\rev{We also train our full pairwise registration module using \cite{mayer2016large} with \emph{learned bases}.}

\begin{figure}[!t]
\centering
\includegraphics[width=\linewidth]{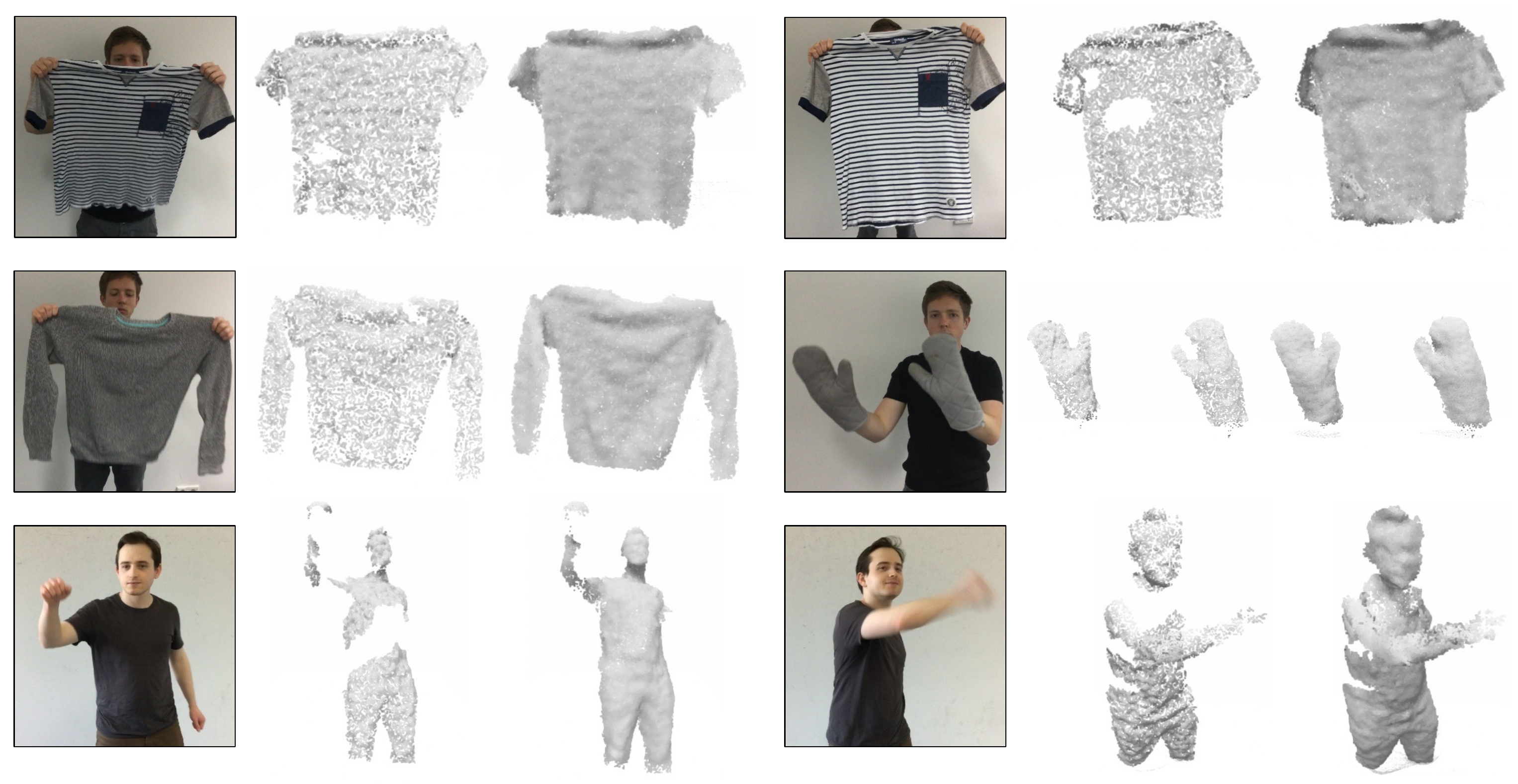}
\vspace{-1.5em}
\caption{\textbf{Dynamic reconstruction on \ddd~dataset} by accumulating the input point clouds. Left: RGB image of the reference frame (\texttt{ref}), which are \emph{not} taken as input. Middle: Unprojected depth point cloud from \texttt{ref}. Right: Accumulated points by warping other 3 frames to \texttt{ref}.}
\label{fig:dd}
\vspace{-1.2em}
\end{figure}

\begin{figure}[!t]
\centering
\includegraphics[width=\linewidth]{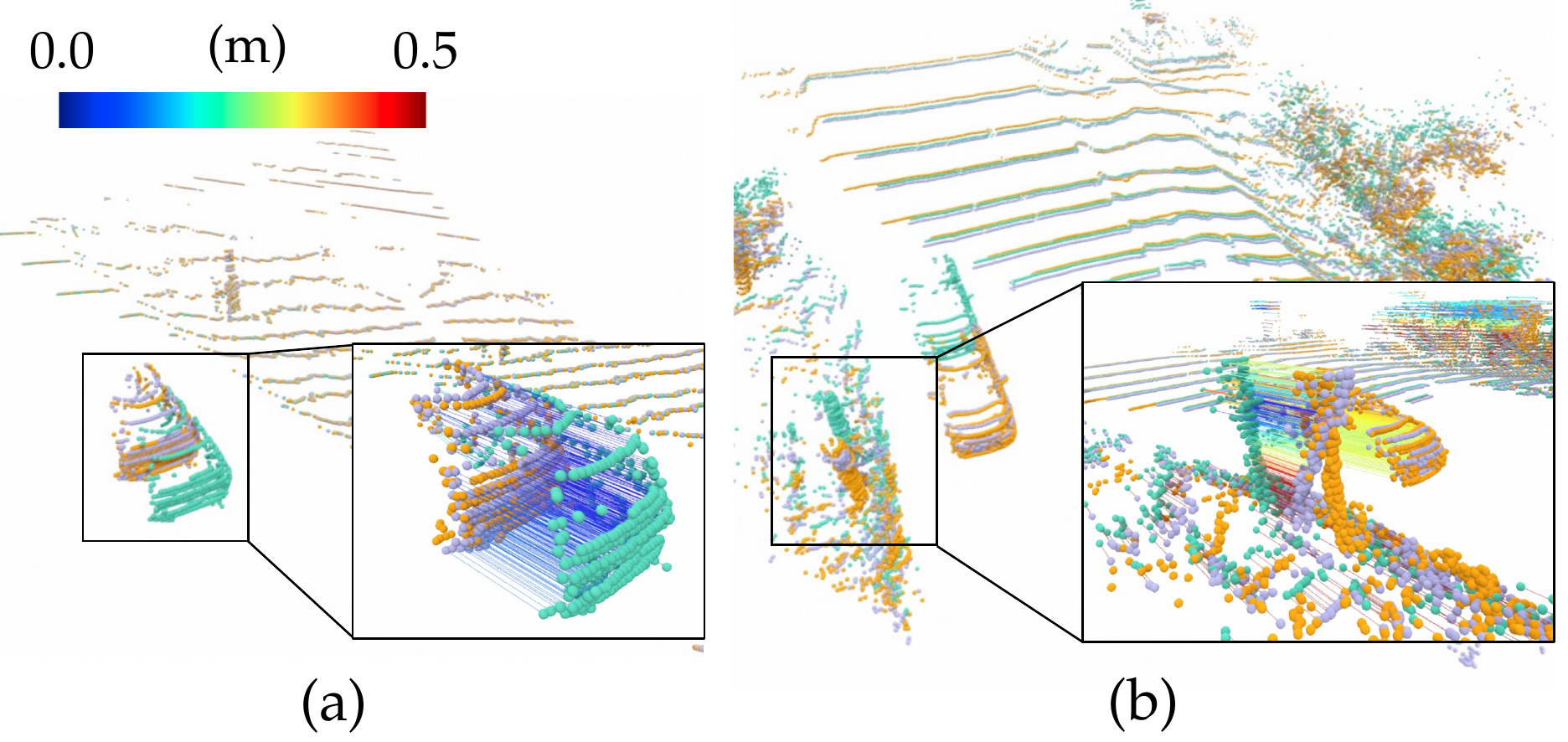}
\caption{\rev{\textbf{Results on LidarKITTI dataset.} (a) and (b) show the best and worst result estimated using our method, respectively (\epe s are 1.7cm and 87.7cm). \cbox{caribbeangreen} Source points; \cbox{orange} Target points; \cbox{amethyst} Warped points.}}
\label{fig:kitti}
\vspace{-1em}
\end{figure}

The flow estimation results are listed in \cref{tbl:kitti}. Note that here the metric ratios use the original thresholds defined in \cite{liu2019flownet3d}.
Compared to our baselines, ours with affinity bases reach a better performance, showing that the basis structure can effectively introduce meaningful priors to the scene flow estimation.
Visualized in \cref{fig:kitti}, the main error comes from the affinely-warped background due to wrong raw correspondences estimated by $\corr$.
In fact, scene flow estimation in the wild achieves improved performance mainly due to more accurate ego-motion estimation as already pointed out in, \eg, \cite{baur2021slim} and our result can be further improved if such rigidity constraint is added. Though, this is out of our current scope, as our aim is to validate the effectiveness of our affinity bases.

\rev{
Nevertheless, we find that in such large-scale multi-object scenes, the learned global bases are less accurate than hand-crafted ones.
This is because in such scenes, each object moves \emph{independently} and each basis should thus be concentrated on a \emph{single} rigid body. However, the number of bodies vary between scenes and the boundaries are hard to fit.
These properties counteract our $\basisnet$ network and are challenging to learn.
Such problems do not exist in object-level datasets where the geometric structures are simpler.
}

\vspace{-1em}
\subsection{Ablation Studies}
\label{subsec:ablation}

\input{tables/sapien-full}

\input{tables/kitti}

\numberedparahead{Number of Bases}
The number of basis functions $M$ reflects the prior knowledge on the intrinsic motion variability of the input shape. 
\cref{tbl:nbasis} shows that as the number of bases increases, the flow estimation error first decreases, but after passing a certain point, rises significantly.
This is because the noise in the descriptors mistakenly incorporated into the learned bases start to dominate the increased representation power brought by adding more bases.
In the meantime, computing the pseudo-inverse of a wider $\basis_k$ matrix demands stronger numerical stability hampering the convergence of the entire model.
In practice, we found that selecting $M=24$ strikes a good balance between accuracy and computation time of the synchronization, which grows exponentially as $M$ increases.

\input{tables/nbasis}

\begin{figure}[!t]
\centering
\includegraphics[width=\linewidth]{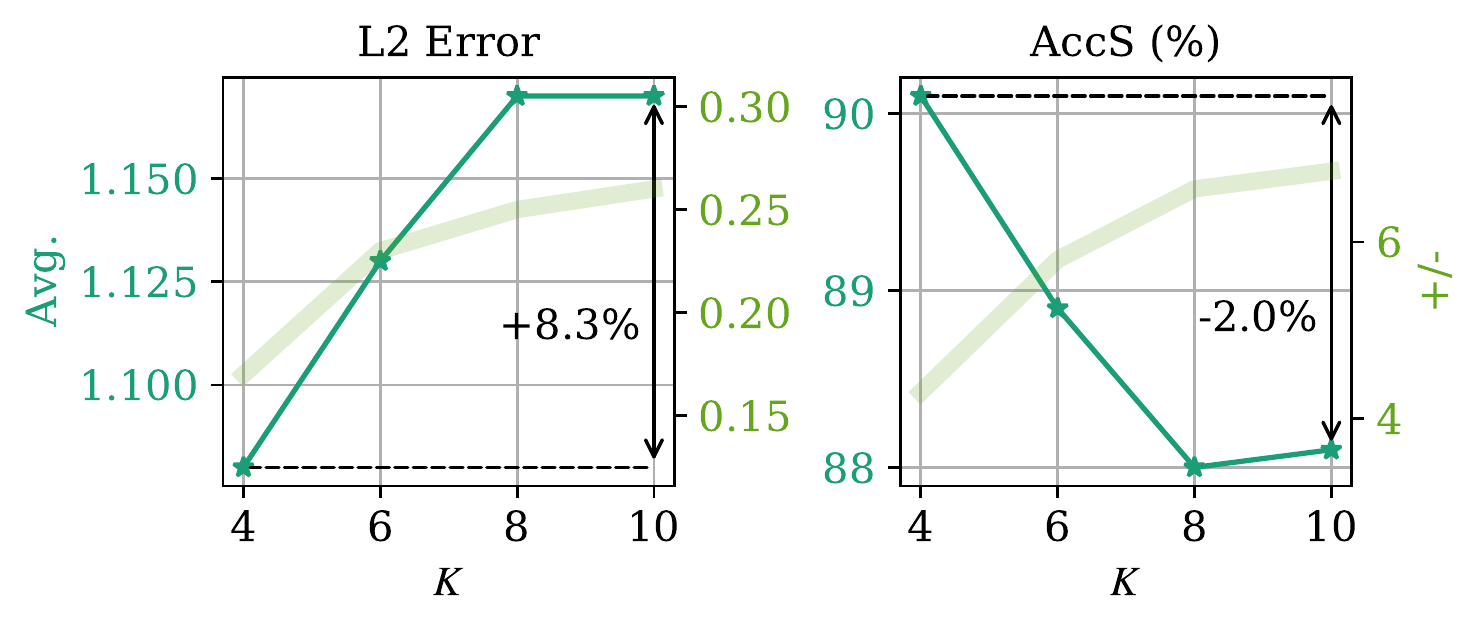}
\vspace{-2.5em}
\caption{\textbf{Influence of frame count $K$.} Changes of \epe{} and AccS \wrt the number of frames $K$ are shown separately in the two plots. Dark green thin curves show the average metric (Avg.) while light green bold curves show the standard deviation (+/-). The black arrows highlight the changes of the average metrics from $K=4$ to $K=10$.}
\label{fig:mf-paper}
\vspace{-1em}
\end{figure}

\numberedparahead{Number of Frames}
\rev{
Our method can be easily applied to more frames ($K>4$) without re-training. 
To demonstrate this, we additionally sample $K=6,8,10$ frames (under the same time interval) for \dcape{} dataset and test our trained model on them.
As shown in \cref{fig:mf-paper}, the accuracy of our method only drops marginally when $K$ increases.
This is because all of our networks are trained in the pairwise setting and there is no learnable component in the multi-frame synchronization module.
}

\numberedparahead{Choice of Basis}
Different basis functions defined on point clouds are available and we compare \name~to one of the non-learned methods~\cite{Sharp:2020:LNT}.
Despite being proven by \cite{aflalo2015optimality} to be \emph{optimal} in representing smooth functions on manifold (in the sense that its gradient magnitudes are bounded), such bases functions are not task-tailored, whereas ours are learned to maximize the scene flow estimation performance and can work robustly under noise.
As shown in \cref{tbl:basis}, even with 80 bases, the flow recovered using \cite{Sharp:2020:LNT} is worse than ours with only 24 bases.
Moreover, the basis computation from a sparse convolution backbone accelerated with GPU is highly efficient and benefits from parallel implementation.
The reduced number of bases not only makes the eigen-decomposition faster, but also boosts the efficiency of subsequent synchronization module.

\numberedparahead{Training Scheme of $\basisnet$}
We validate the effectiveness of the training strategies applied for $\basisnet$, including (a) the use of the robust term $\rho(\cdot)$, (b) the addition of the consistency loss $\mathcal{L}_\mathrm{c}$ and (c) the mixing strategy of $\corr_\gt$ and $\corr$.
As demonstrated in \cref{tbl:bnet-train}, adding the robust term gives the framework more tolerance towards noise while the consistency loss acts as an efficient regularizer and speeds up network convergence.
Apart from the performance gain from the mixing strategy, we empirically found that using only the ground-truth $\corr$ causes singularities in the network prediction which gives ill-posed scene flow when two frames are too close.

\numberedparahead{Refinement Strategy}
We show that the combination of flows in \cref{eq:flowfin} is indeed helpful, especially in the occluded regions.
As illustrated in \cref{tbl:refine}, feeding the three pieces of information altogether helps the refinement module achieve the best result.
\rev{$\X$ acts as a regularizer by reducing gross errors. Despite its decisive effect on flow accuracy, $\flow^\nn$ is, by definition, not capable of transferring points to the empty region, where $\flow^\bs$ comes to the rescue by extending the target mapped positions to the full 3D space} (\cf \cref{fig:occlusion} for visual comparisons).

\input{tables/basis}

\input{tables/bnet-train.tex}

\input{tables/refine}

\numberedparahead{Generalization}
\label{para:general}
\rev{%
To further demonstrate how our model generalizes to unseen domains, we re-split the \emph{animals} part of the \ddtfd{} dataset into two \emph{disjoint} category sets $\Cset_1$ and $\Cset_2$ (detailed in the appendix).
Results in \cref{tbl:general} show that our model trained only on $\Cset_1$ can reach a similar performance to the model trained on $\Cset_1 \cup \Cset_2$ when tested on $\Cset_2$ (note the \emph{actions} of the training and test set never overlap, denoted by $\Aset_1, \Aset_2, \Aset_3$).
This empirically verifies that our method can generalize across different animal breeds.
However, our method \emph{fails} to generalize from \emph{humanoids} (denoted as $\Cset_\human$) to novel categories with drastically different geometries like $\Cset_2$. As echoed in \cref{sec:conclude}, this probably stems from the fact that our descriptor network and basis network overfit to the global structural information.
}

\input{tables/general}

\begin{figure}[!t]
\centering
\includegraphics[width=\linewidth]{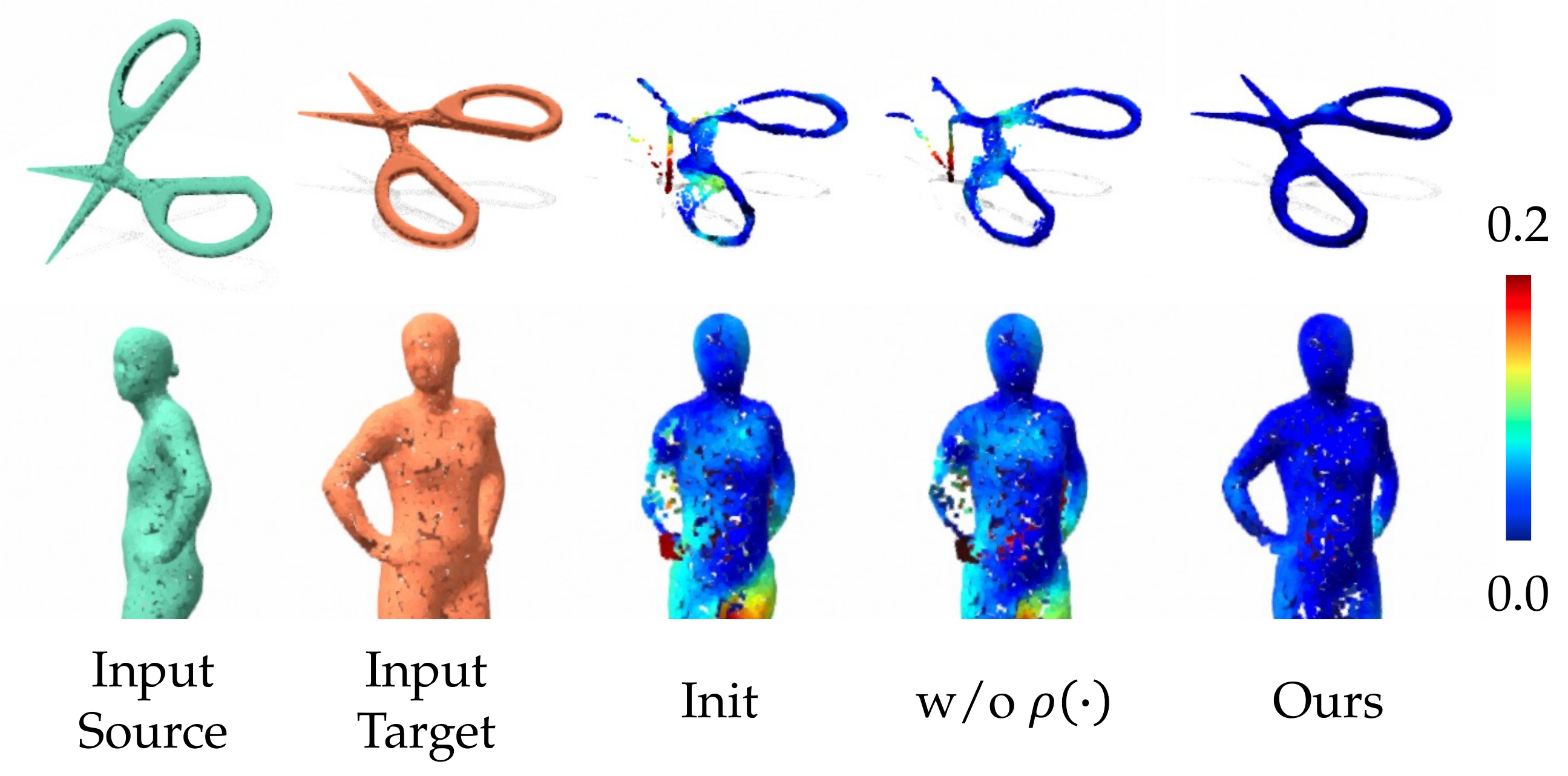}
\vspace{-2em}
\caption{\textbf{Visual comparison of the synchronization effect.} Flow error is shown on the warped source point cloud. The last 2 columns are computed using additional inputs (not shown here) for synchronization.}
\label{fig:worobust}
\vspace{-1em}
\end{figure}

\numberedparahead{Effect of Synchronization}
Synchronization links the information from multiple input point clouds exploiting cycle consistency and reduces the overall error.
Here, we ablate different optimization strategies in \cref{tbl:worobust} using \dsapien~dataset.
The use of the robust term \underline{$\rho(\cdot)$}, different from the one used for pairwise training, now jointly considers multi-scan cues and prunes the inconsistent outliers which is hard to detect in the pairwise setting.
Although the basis preconditioning (\underline{Basis Precond.}) strategy does not result in a large performance difference at convergence, it has a positive impact on the convergence rate.
As a direct replacement to our alternating optimization scheme (\underline{Alter. Optim.}), we borrow the solver from \cite{pymanopt} to jointly optimize $\Hfunc$ and $\C$ on the product manifold of Stiefel and Euclidean.
At convergence, this direct optimizer reaches a slightly lower error. 
Nevertheless, a lower flow error does not necessarily indicate a better convergence of \cref{eq:sync} and vice versa.
We conclude, though, that our optimization scheme can reach a \emph{decent performance} with \emph{much better efficiency} than its counterparts.
Please see \cref{fig:worobust} for visual comparisons.

\begin{figure}[!htbp]
\centering
\includegraphics[width=\linewidth]{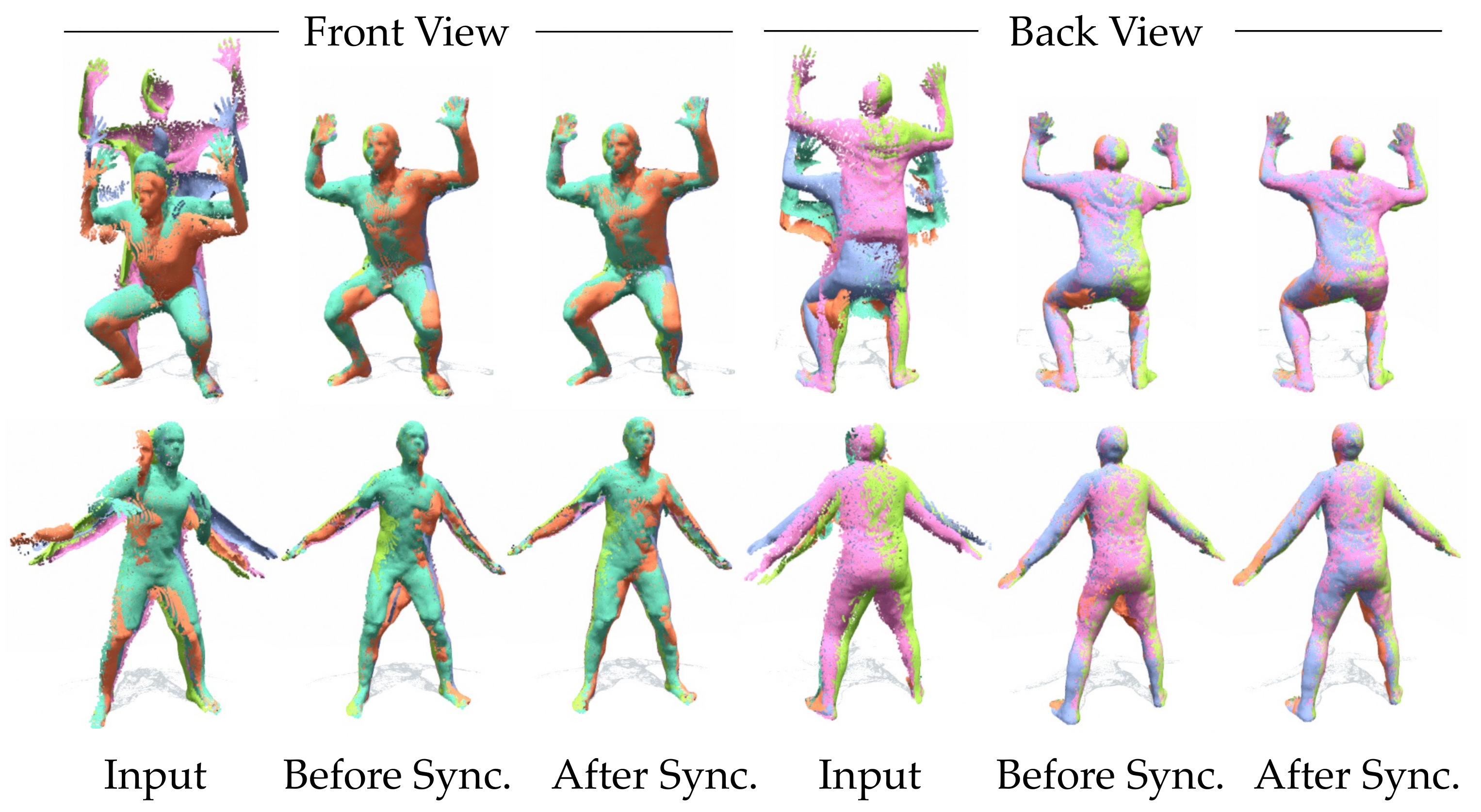}
\vspace{-2.0em}
\caption{\rev{\textbf{Qualitative results on \dcape{}-360.} \emph{Input} column shows overlayed point clouds from all 5 views (different colors denote different views), while \emph{Before Sync.} and \emph{After Sync.} show the warped point clouds obtained from our full pipeline, before and after synchronization, respectively.}}
\label{fig:rot-human}
\end{figure}

\rev{Furthermore, to show the effect under large view differences, we create a \dcape{}-360 dataset, where a moving camera is circling around the performing human actor, and captures 5 frames in between. All individual views are partial, but combined, they cover the full body. After an additional post-processing step of rigidly aligning the point clouds from camera space to world space, our method can successfully align different views and assemble a complete body, as illustrated in \cref{fig:rot-human}. The synchronization module closes the loop by jointly considering multiple views and corrects the alignment errors, which are hard to tackle under the pairwise setting.}

\input{tables/worobust}

\vspace{-0.5em}
\subsection{Timing and Memory Analysis}

\includegraphics[width=\linewidth]{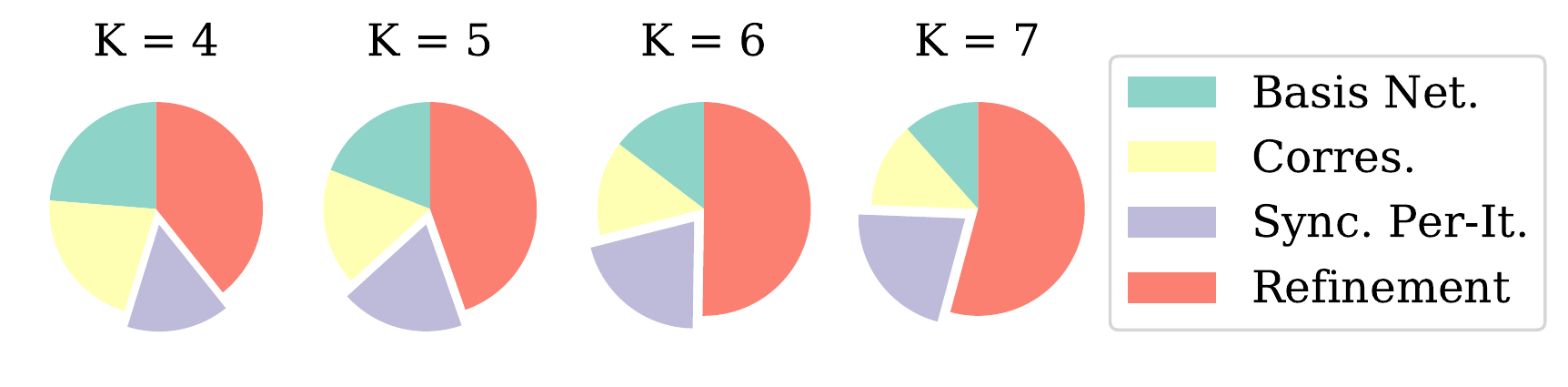}

We evaluate the efficiency of \name~on a workstation with a 3.70GHz Intel i9 CPU with a single GeForce RTX 3090 graphics card.
On average it takes 1.22s, 1.86s, 2.80s, 3.89s to generate 12, 20, 30, 42 pairs of flow vectors for $K=4, 5, 6, 7$ (\ie, roughly 100ms/pair), respectively. 
The detailed break-down of the relative time cost by each component is shown in the pie chart above: 
while the refinement module (\cbox{set3_red}) takes more time than $\basisnet$ (\cbox{set3_green}) and computing $\corr$ (\cbox{set3_yellow}) when $K$ gets larger (because the number of pairs grow quadratically), the synchronization module (\cbox{set3_purple}) is always the one that dominates the computation.
We find that the training of both $\basisnet$ and $\refnet$ usually reaches a full converge within two days, while baselines like PWC take more than 5 days.
The parameters for $\basisnet$, $\descnet$ and $\refnet$ are 8.8M, 5.0M and 0.7M, respectively.
Runtime memory footprint for the entire pipeline is around 2.8GiB for $K=4$.

\begin{figure}[!t]
\centering
\includegraphics[width=\linewidth]{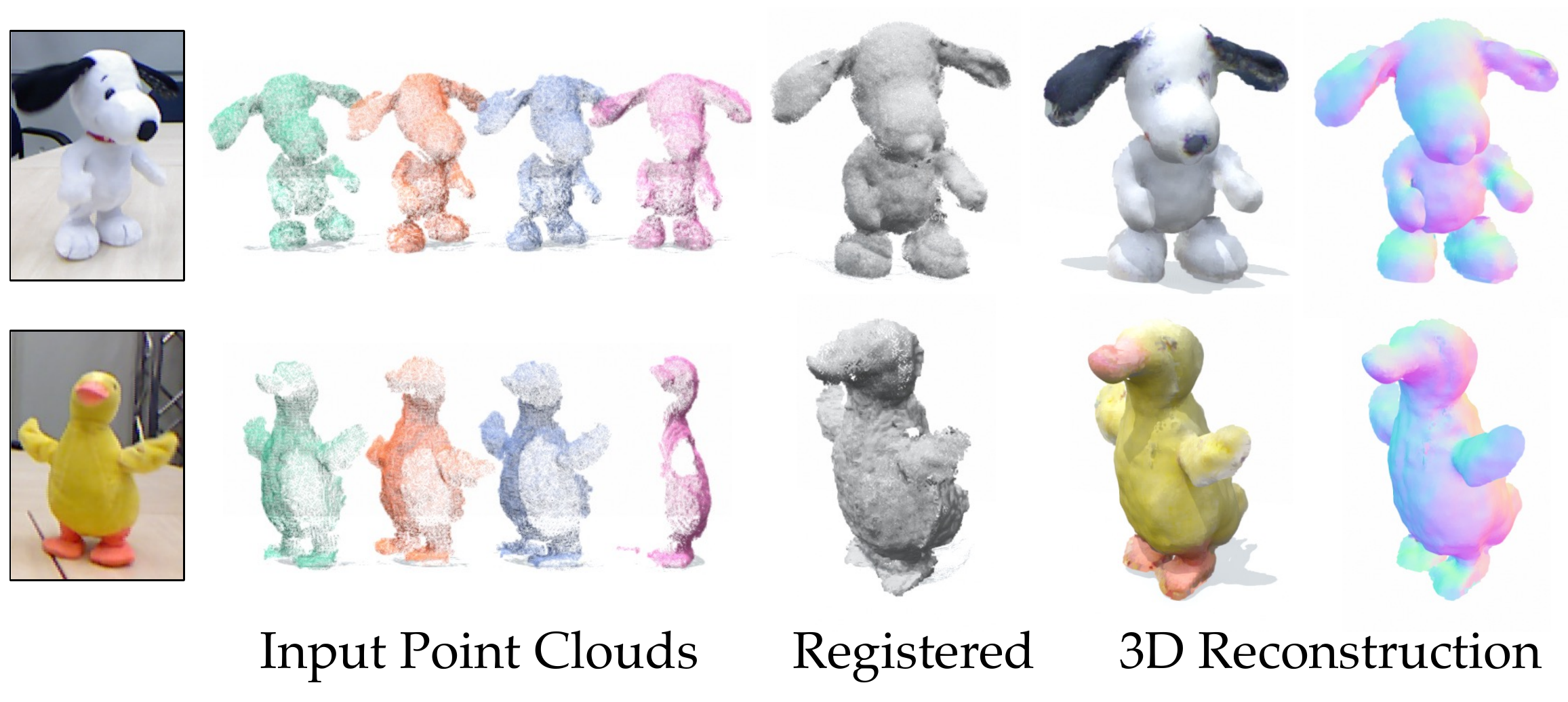}
\vspace{-2em}
\caption{\rev{\textbf{Non-rigid reconstruction.} We show demonstrative results on the dataset from \cite{slavcheva2017killingfusion}. Reconstructions are obtained via \cite{kazhdan2013screened}.}}
\label{fig:slav}
\vspace{-1em}
\end{figure}

\begin{figure}[!t]
\centering
\includegraphics[width=\linewidth]{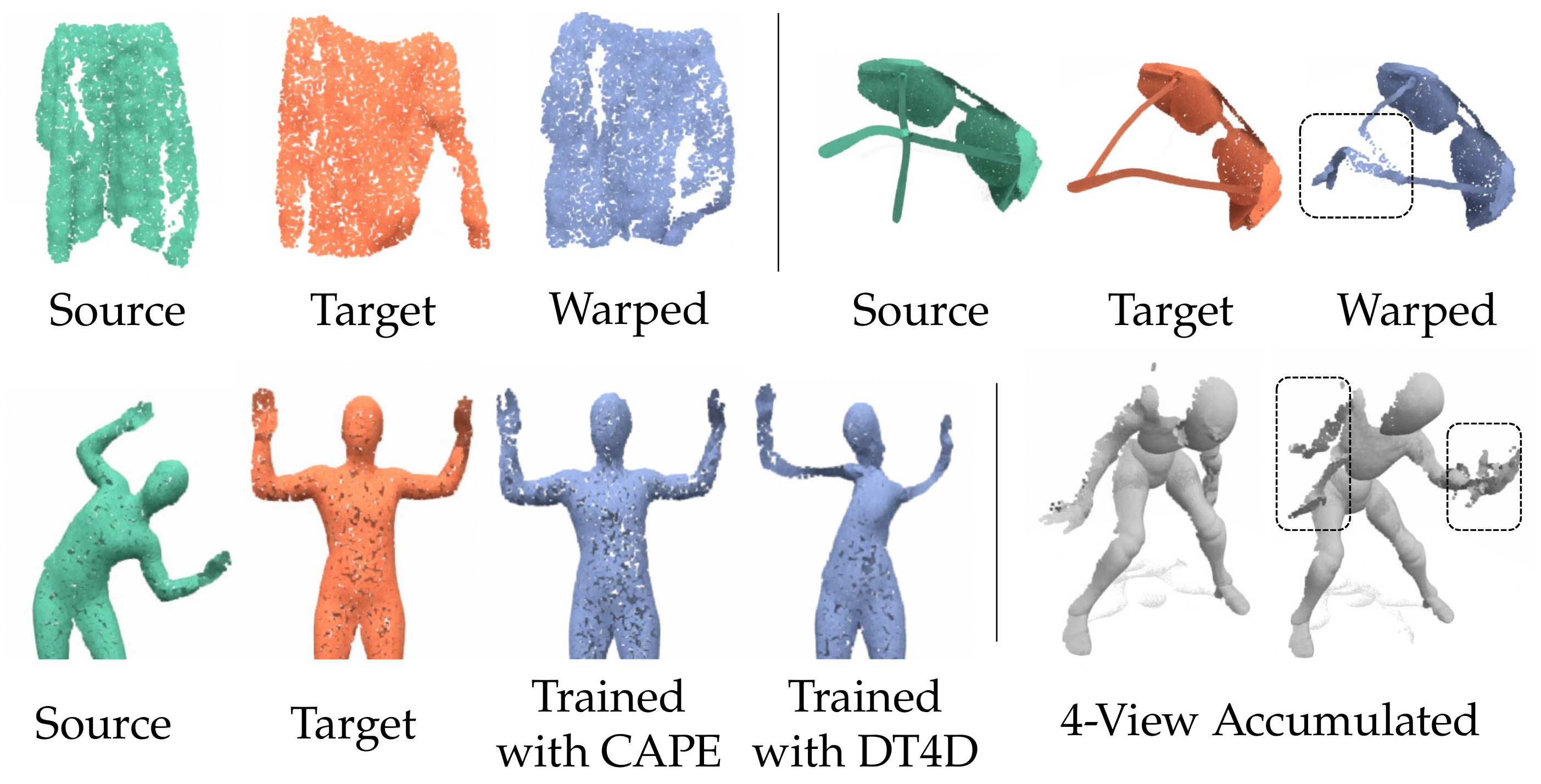}
\vspace{-2em}
\caption{\textbf{Failure Cases}. 
Large topological changes and matching ambiguities fail \name, suffering limited cross-domain generalizability.}
\label{fig:failure}
\vspace{-1.2em}
\end{figure}

%% file: tables/dataset.tex
\begin{table}[!t]
\setlength{\tabcolsep}{2.8pt}
\caption{\textbf{Dataset statistics.} We report the number of training, validation and test samples used in the quantitative evalutions.}
\label{tbl:dataset}
\begin{center}
\footnotesize
\vspace{-1.5em}
\begin{tabular}{l|ccccc}
\toprule
Dataset           & \# Train / Val. & \# Test & Partial & Non-rigid & Real \\ \midrule
\dcape~\cite{cape1,cape2}              & 3015 / 798           & 209           &    -    & \checkmark         & \checkmark$^\dagger$          \\
\ddtfd~\cite{Li20214DCompleteNM} & 3907 / 1701          & 1299          & \checkmark       & \checkmark         &     -      \\
\ddd~\cite{Bozic2020DeepDeformLN}        & 1754 / 200           & 267           & \checkmark       & \checkmark         & \checkmark          \\
\dsapien~\cite{Xiang_2020_SAPIEN}            & 530 / 88             & 266           &  -      &   -       & -          \\ \bottomrule
\end{tabular}
\end{center}
\vspace{-0.5em}
$^\dagger$: For train/val we use the points sampled from the parametric meshes while for test we use the raw-scanned point clouds.
\vspace{-1em}
\end{table}

%% file: tables/cape-full.tex
\begin{table*}[!t]
\setlength{\tabcolsep}{2.1em}
\caption{\textbf{Quantitative comparison on \dcape~dataset} using fully-supervised and self-supervised schemes. $\downarrow / \uparrow$: Lower/higher is better (for the standard deviation being lower is always better). Supervision with `Mesh' requires an additional template (or parametric) human mesh as a canonical prior. `w/o sync.' is inferred from the initial functional map matrices without synchronization. \textbf{Boldfaced} numbers highlight the best and \underline{underlined} numbers mark the tied first. Run time in (parentheses) denotes no GPU acceleration.}
\label{tbl:cape-full}
\begin{center}
\footnotesize
\vspace{-1.5em}
\begin{tabular}{@{}l|c|ccccc@{}}
\toprule
                   & Supervision & \epe~$\downarrow$       & AccS $\uparrow$       & AccR $\uparrow$       & Outlier $\downarrow$    & Run Time (s) \\ \midrule
CPD~\cite{myronenko2010cpd}                & -           & 7.06 $\pm$ 2.46  & 2.5 $\pm$ 2.1   & 42.0 $\pm$ 20.3 & 90.8 $\pm$ 5.5  &    1.03      \\
GUIP~\cite{Giraldo2017GroupWisePR} & -           & 5.55 $\pm$ 2.16  & 35.8 $\pm$ 11.1 & 69.3 $\pm$ 13.4 & 80.9 $\pm$ 6.5  &   (28.2)   \\
NPP~\cite{hayden2020nonparametric}                & -           & 19.02 $\pm$ 5.63 & 1.3 $\pm$ 1.7   & 10.6 $\pm$ 8.7  & 99.3 $\pm$ 0.9  & (70.0)    \\ \midrule
CZO~\cite{huang2020consistent}                & Full       & 2.04 $\pm$ 0.27  & 64.2 $\pm$ 7.5  & 94.2 $\pm$ 1.9  & 51.1 $\pm$ 10.7 &   (24.2)  \\
3D-CODED~\cite{groueix20183d}      & Full + Mesh      & 2.25 $\pm$ 0.58  & 61.6 $\pm$ 9.0  & 95.4 $\pm$ 3.6  & 53.8 $\pm$ 11.2 &   160$^\ddagger$  \\
PTF~\cite{wang2021locally}                & Full + Mesh       & 2.52 $\pm$ 0.54  & 47.9 $\pm$ 12.8  & 91.4 $\pm$ 5.7  & 57.8 $\pm$ 9.6 &   100  \\
PWC~\cite{wu2019pointpwc}                & Full        & 1.37 $\pm$ 0.34  & 82.2 $\pm$ 8.4  & 98.3 $\pm$ 1.6  & 39.1 $\pm$ 10.4 &   0.99  \\
FLOT~\cite{puy20flot}               & Full        & 1.79 $\pm$ 0.34  & 69.9 $\pm$ 9.3  & 96.9 $\pm$ 2.0  & 48.4 $\pm$ 10.7 &   1.45   \\
\rev{EmbAlign}~\cite{marin2020correspondence}  & Full        & 2.15 $\pm$ 0.49  & 61.5 $\pm$ 10.6  & 95.0 $\pm$ 3.5  & 53.3 $\pm$ 10.0 &   0.60   \\
\rev{GeomFMaps}~\cite{donati2020deep}  & Full        & 2.52 $\pm$ 0.25  & 54.3 $\pm$ 6.2  & 88.1 $\pm$ 2.0  & 54.4 $\pm$ 10.7 &   1.40   \\
MBS~\cite{huang2021multibodysync}               & Full        & 2.15 $\pm$ 0.22  & 54.0 $\pm$ 5.4  & 97.2 $\pm$ 1.8  & 55.9 $\pm$ 11.4 &  1.12   \\
\textbf{Ours} {\scriptsize($\descnet$ only)}  & Full        & 1.39 $\pm$ 0.18  & 82.0 $\pm$ 5.3  & 99.1 $\pm$ 0.6 & 42.4 $\pm$ 10.9 &  0.12   \\
\textbf{Ours}               & Full        & \textbf{1.08} $\pm$ \textbf{0.17}  & \textbf{90.1} $\pm$ \textbf{4.3}  & \textbf{99.5} $\pm$ \textbf{0.5}  & \textbf{35.3} $\pm$ \textbf{10.1} &  1.22   \\ \midrule
PWC-U~\cite{wu2019pointpwc}                & Self        & 3.52 $\pm$ 1.68  & 50.0 $\pm$ 13.6 & 82.7 $\pm$ 11.7 & 67.4 $\pm$ 8.7  & 0.99    \\
RMA~\cite{feng2021recurrent}           & Self$^\dagger$        & 5.47 $\pm$ 2.29  & 29.5 $\pm$ \underline{15.0} & 67.9 $\pm$ 15.5 & 83.4 $\pm$ 7.0  &  1.60    \\
\rev{DeepShells}~\cite{eisenberger2020deep}                & Self        & 4.55 $\pm$ 3.61  & 57.1 $\pm$ 21.6 & 81.2 $\pm$ 20.5 & 60.6 $\pm$ 16.4  & 58.0    \\
\textbf{Ours} {\scriptsize($\descnet$ only)}        & Self        & 4.20 $\pm$ 1.44  & 34.4 $\pm$ 9.5 & 79.1 $\pm$ 10.0  & 71.5 $\pm$ 8.5 &  0.12        \\
\textbf{Ours} \wosync        & Self        & 3.07 $\pm$ 1.37  & 53.9 $\pm$ 15.3 & 88.1 $\pm$ 8.7  & 60.3 $\pm$ \underline{11.0} &  0.32        \\
\textbf{Ours}   & Self        & \textbf{2.95} $\pm$ \textbf{1.29}  & \textbf{55.5} $\pm$ \underline{15.0} & \textbf{89.1} $\pm$ \textbf{8.0}  & \textbf{59.1} $\pm$ \underline{11.0} &  1.22        \\ \bottomrule
\end{tabular}
\end{center}
\vspace{-0.5em}
$^\dagger$: Fails to converge on our dataset if trained from scratch so we finetune from the pretrained model provided by the official repository.
\\$^\ddagger$: Includes the time for best rotation selection and test-time Chamfer loss minimization.
\end{table*}

%% file: tables/faust.tex
\begin{table*}[!ht]
\caption{\textbf{FAUST intra-subject challenge~\cite{bogo2014faust} results (Top9)}. Methods are ordered from left to right according to their rankings.} 
\label{tbl:faust}
\vspace{-2em}
\begin{center}
\footnotesize
\begin{tabular}{@{}c|ccccccccc@{}}
\toprule
           & DHNN~\cite{jiang2020disentangled} & DVM~\cite{kim2021deep}   & \textbf{Ours}  & SP~\cite{zuffi2015stitched} & AtlasNetV2~\cite{deprelle2019learning} & 3D-CODED~\cite{groueix20183d} & BPS~\cite{prokudin2019efficient} & LBS-AE~\cite{li2019lbs} & FMNet~\cite{litany2017deep} \\ \midrule
Error (cm) & 1.011        & 1.185$^\dagger$ / 1.417 & \underline{1.486} & 1.568      & 1.626              & 1.985            & 2.010       & 2.161          & 2.436         \\ \bottomrule
\end{tabular}
\end{center}
\vspace{-0.5em}
$^\dagger$: Requires multi-view (72) input that requires multiple forward passes of the model. Costs $\sim$7s/pair.
\vspace{-1.5em}
\end{table*}

%% file: tables/dt4d-full.tex
\begin{table*}[!t]
\setlength{\tabcolsep}{1.05em}
\caption{\textbf{Quantitative comparison on \ddtfd~dataset.} We report results for both the non-occluded set of points and the full point cloud. \\ See caption of \cref{tbl:cape-full} for other legends.}
\label{tbl:dt4d-full}
\centering
\footnotesize
\vspace{-1.5em}
\begin{tabular}{@{}l|cccc|cccc@{}}
\toprule
\multirow{2}{*}{} & \multicolumn{4}{c|}{\emph{Non-Occluded}}                      & \multicolumn{4}{c}{\emph{Full}}                               \\
                  & \epe~$\downarrow$       & AccS $\uparrow$       & AccR $\uparrow$       & Outlier $\downarrow$    & \epe~$\downarrow$       & AccS $\uparrow$       & AccR $\uparrow$       & Outlier $\downarrow$    \\ \midrule
CPD~\cite{myronenko2010cpd}               & 10.12 $\pm$ 4.28 & 11.7 $\pm$ \underline{8.9}  & 43.2 $\pm$ 18.1 & 67.3 $\pm$ 12.8 & 10.91 $\pm$ 4.69 & 11.4 $\pm$ \textbf{8.6}  & 41.8 $\pm$ 17.8 & 68.4 $\pm$ 12.3 \\
GUIP~\cite{Giraldo2017GroupWisePR}             & 16.43 $\pm$ 7.42 & 20.3 $\pm$ \underline{8.9}  & 43.0 $\pm$ 13.5 & 73.6 $\pm$ 10.6 & 17.22 $\pm$ 7.66 & 19.4 $\pm$ 8.8  & 41.2 $\pm$ 13.5 & 74.6 $\pm$ 10.2 \\
CZO~\cite{huang2020consistent}               & 4.93 $\pm$ 1.62  & 42.4 $\pm$ 12.1 & 75.2 $\pm$ 10.3 & 37.8 $\pm$ 11.3 & 6.44 $\pm$ 2.45  & 39.8 $\pm$ 11.9 & 70.7 $\pm$ 11.2 & 40.8 $\pm$ 11.0 \\
PWC~\cite{wu2019pointpwc}               & 5.39 $\pm$ 2.84  & 47.5 $\pm$ 14.0 & 77.0 $\pm$ 12.2 & 36.8 $\pm$ 12.5 & 6.35 $\pm$ 3.43  & 44.9 $\pm$ 13.9 & 73.2 $\pm$ 13.1 & 39.3 $\pm$ 12.5 \\
FLOT~\cite{puy20flot}              & 5.14 $\pm$ 1.94  & 43.2 $\pm$ 11.7 & 75.3 $\pm$ 10.0 & 39.8 $\pm$ 10.9 & 6.57 $\pm$ 2.92  & 40.3 $\pm$ 11.6 & 70.5 $\pm$ 11.2 & 43.0 $\pm$ 10.9 \\
\rev{EmbAlign}~\cite{marin2020correspondence}         & 6.97 $\pm$ 2.75  & 36.3 $\pm$ 11.4 & 66.1 $\pm$ 11.3 & 48.1 $\pm$ 10.5 & 8.22 $\pm$ 3.56  & 34.1 $\pm$ 11.3 & 62.3 $\pm$ 12.1 & 50.7 $\pm$ 10.4 \\
MBS~\cite{huang2021multibodysync}               & 5.55 $\pm$ 2.45  & 43.9 $\pm$ 11.5 & 76.3 $\pm$ 11.1 & 38.4 $\pm$ 11.7 & 6.70 $\pm$ 3.13  & 41.2 $\pm$ 11.6 & 72.1 $\pm$ 12.1 & 41.1 $\pm$ 11.8 \\ \midrule
\textbf{Ours} {\scriptsize($\descnet$ only)}       & 3.77 $\pm$ 1.54  & 63.0 $\pm$ 9.8 & 86.5 $\pm$ 6.0  & 28.2 $\pm$ 9.9 & 5.40 $\pm$ 2.78  & 58.2 $\pm$ 10.6 & 81.2 $\pm$ 8.4  & 31.9 $\pm$ 10.3 \\
\textbf{Ours} \wosync       & 3.04 $\pm$ 1.54  & \underline{69.2} $\pm$ 10.7 & 89.5 $\pm$ 6.4  & 24.1 $\pm$ 10.1 & 3.73 $\pm$ 2.05  & \underline{65.5} $\pm$ 11.5 & 86.3 $\pm$ 8.0  & 26.0 $\pm$ 10.4 \\
\textbf{Ours}       & \textbf{2.87} $\pm$ \textbf{1.25}  & \underline{69.2} $\pm$ 10.2 & \textbf{89.8} $\pm$ \textbf{5.7}  & \textbf{23.8} $\pm$ \textbf{9.7}  & \textbf{3.53} $\pm$ \textbf{1.71}  & \underline{65.5} $\pm$ 10.9 & \textbf{86.6} $\pm$ \textbf{7.2}  & \textbf{25.6} $\pm$ \textbf{9.9}  \\ \bottomrule
\end{tabular}
\vspace{-1em}
\end{table*}

%% file: tables/sapien-full.tex
\begin{table}[!t]
\setlength{\tabcolsep}{4.0pt}
\caption{\textbf{Quantitative comparison on \dsapien~dataset.} \\ See caption of \cref{tbl:cape-full} for other legends.}
\label{tbl:sapien-full}
\centering
\footnotesize
\vspace{-1em}
\begin{tabular}{@{}l|cccc@{}}
\toprule
                   & \epe~$\downarrow$       & AccS $\uparrow$       & AccR $\uparrow$       & Outlier $\downarrow$    \\ \midrule
CPD~\cite{myronenko2010cpd}                & 9.18 $\pm$ 4.50  & 7.4 $\pm$ \textbf{5.8}   & 37.1 $\pm$ 22.7 & 81.3 $\pm$ 13.0 \\
GUIP~\cite{Giraldo2017GroupWisePR}            & 10.85 $\pm$ 4.28 & 10.0 $\pm$ 8.0  & 30.4 $\pm$ 16.8 & 84.7 $\pm$ \textbf{6.9}  \\
CZO~\cite{huang2020consistent}               & 3.40 $\pm$ 2.12  & 62.3 $\pm$ 14.5 & 86.9 $\pm$ 12.4 & 24.5 $\pm$ 11.8 \\
PWC~\cite{wu2019pointpwc}        & 8.95 $\pm$ 3.86  & 7.3 $\pm$ 7.0   & 35.6 $\pm$ 20.9 & 81.3 $\pm$ 11.9 \\
FLOT~\cite{puy20flot}      & 5.17 $\pm$ 2.69  & 32.9 $\pm$ 16.0 & 71.3 $\pm$ 18.6 & 45.6 $\pm$ 11.9 \\
\rev{EmbAlign}~\cite{marin2020correspondence} & 11.12 $\pm$ 4.94  & 7.1 $\pm$ 6.1 & 28.2 $\pm$ 17.7 & 88.6 $\pm$ 6.0 \\
MBS~\cite{huang2021multibodysync}          & 4.93 $\pm$ 2.27  & 34.8 $\pm$ 14.4 & 72.8 $\pm$ 16.2 & 43.6 $\pm$ 10.5 \\ \midrule
\textbf{Ours} {\scriptsize($\descnet$ only)}        & 4.66 $\pm$ 2.45  & 38.4 $\pm$ 13.0 & 78.4 $\pm$ 13.9 & 39.1 $\pm$ 9.7 \\
\textbf{Ours} \wosync        & 3.48 $\pm$ 2.41  & 61.0 $\pm$ 19.0 & 86.2 $\pm$ 13.2 & 25.1 $\pm$ 11.9 \\
\textbf{Ours}  & \textbf{3.09} $\pm$ \textbf{1.68}  & \textbf{62.8} $\pm$ 15.3 & \textbf{88.1} $\pm$ \textbf{9.7}  & \textbf{23.7} $\pm$ 9.6  \\ \bottomrule
\end{tabular}
\vspace{-1em}
\end{table}

%% file: tables/kitti.tex
\begin{table}[!t]
\setlength{\tabcolsep}{3.0pt}
\caption{\textbf{Comparisons on LidarKITTI dataset.} Rigid3DSceneFlow~\cite{Gojcic2021WeaklySL} exploits explicit rigid ego-motion estimation for the background scene flow. ($\descnet$ only) simply maps each point to its nearest neighbor in the feature space embedded by $\descnet$.}
\label{tbl:kitti}
\centering
\footnotesize
\vspace{-1em}
\begin{tabular}{@{}l|cccc@{}}
\toprule
                  & EPE3D $\downarrow$ & AccS $\uparrow$ & AccR $\uparrow$ & Outliers $\downarrow$ \\ \midrule
PointPWC-Net~\cite{wu2019pointpwc}      & 39.0       & \textbf{38.7}        & 55.0        & \textbf{65.3}          \\
FLOT~\cite{puy20flot}              & 65.3       & 15.5        & 31.3        & 83.7          \\ \midrule
\rev{\textbf{Ours} {\scriptsize($\descnet$ only)}} & 57.0       & 31.3        & 45.6        & 73.4          \\
\rev{\textbf{Ours} {\scriptsize(learned bases)}}             & 46.2       & 38.1        & \textbf{59.6}        & 66.5          \\ 
\rev{\textbf{Ours} {\scriptsize(affinity bases)}}             & \textbf{31.5}       & 34.8        & 55.2        & 70.4          \\ \midrule
Rigid3DSceneFlow~\cite{Gojcic2021WeaklySL}  & \textbf{15.3}       & \textbf{48.5}        & \textbf{73.0}        & \textbf{46.6}          \\ \bottomrule
\end{tabular}
\vspace{-1.0em}
\end{table}

%% file: tables/nbasis.tex
\begin{table}[!t]
\setlength{\tabcolsep}{2em}
\caption{\textbf{Performance using different number of bases} on \dcape~dataset. For run time we measure the average time for each iteration of our synchronization algorithm, in the unit of seconds.}
\label{tbl:nbasis}
\centering
\vspace{-1em}
\footnotesize
\begin{tabular}{c|cc}
\toprule
\# Basis $M$ & \epe~$\downarrow$ & Run Time (sec./iter.) \\ \midrule
8       & 1.14  & $6.0 \times 10^{-3}$       \\
16      & 1.11  & $1.2 \times 10^{-2}$       \\
\textbf{24}      & 1.08  & $2.6 \times 10^{-2}$       \\
32      & 1.05  & $7.0 \times 10^{-2}$       \\
64      & 1.27  & $1.9 \times 10^{0}$         \\
80      & 1.55  & $5.9 \times 10^{0}$         \\ \bottomrule
\end{tabular}
\vspace{-1em}
\end{table}

%% file: tables/basis.tex
\begin{table}[!t]
\caption{\textbf{Comparison of our learned bases with \cite{Sharp:2020:LNT}} on \dcape~dataset. Run time in (parentheses) denotes no GPU acceleration.}
\label{tbl:basis}
\centering
\vspace{-1em}
\footnotesize
\begin{tabular}{@{}l|c|cc|cc@{}}
\toprule
                                        & \multirow{2}{*}{\# Basis} & \multirow{2}{*}{\epe~$\downarrow$} & \multirow{2}{*}{AccS $\uparrow$} & \multicolumn{2}{c}{Run Time (sec.)} \\
                                        &                          &                        &                       & Basis       & Sync. per iter.      \\ \midrule
\multirow{2}{*}{\cite{Sharp:2020:LNT}} & 24                       & 2.96 $\pm$ 0.42            & 48.4 $\pm$ 8.2            & (1.6)         & \textbf{0.03}           \\
             & 80                       & 1.53 $\pm$ 0.21            & 78.4 $\pm$ 5.1            & (3.2)         & 5.9            \\ \midrule
\textbf{Ours}                                    & 24                       & \textbf{1.08} $\pm$ \textbf{0.17}            & \textbf{90.1} $\pm$ \textbf{4.3}            & \textbf{0.24}        & \textbf{0.03}           \\ \bottomrule
\end{tabular}
\end{table}

%% file: tables/bnet-train.tex
\begin{table}[!t]
\setlength{\tabcolsep}{7.0pt}
\caption{\textbf{Comparisons of different $\basisnet$ training strategies} on \ddtfd~dataset. See text for the explanation of (a),(b),(c). The values of EPE3D are measured on (non-occluded / full) point clouds.}
\label{tbl:bnet-train}
\centering
\footnotesize
\vspace{-1em}
\begin{tabular}{@{}ccc|cc@{}}
\toprule
(a) $\rho(\cdot)$ & (b) $\mathcal{L}_\mathrm{c}$ & (c) Mixed $\corr$ & \epe~$\flow^{\nn} \downarrow$  & \epe~$\flow^{\bs} \downarrow$  \\ \midrule
            &            &                          & 3.64 / 5.62 & 4.33 / 6.05 \\
\checkmark  &            &                          & 3.58 / 5.48 & 4.13 / 5.78 \\
\checkmark  & \checkmark &                          & 3.52 / 5.41 & 4.16 / 5.78 \\
\checkmark  &            & \checkmark               & 3.49 / 5.38 & 4.11 / 5.74 \\
            & \checkmark & \checkmark               & 3.23 / 4.76 & 4.19 / 5.49 \\
\checkmark  & \checkmark & \checkmark               & \textbf{3.14} / \textbf{4.59} & \textbf{3.86} / \textbf{5.09} \\ \bottomrule
\end{tabular}
\end{table}

%% file: tables/refine.tex
\begin{table}[!t]
\setlength{\tabcolsep}{15.0pt}
\caption{\rev{\textbf{Comparisons of different refinement strategies} on \ddtfd \\ dataset to generate the final 3D scene flow. \epe s are measured on (non-occluded / full) point clouds.}}
\label{tbl:refine}
\centering
\footnotesize
\vspace{-1em}
\begin{tabular}{@{}cccc@{}}
\toprule
Input + $\flow^\nn$ & Input + $\flow^\bs$ & Input + $\X$ & \epe~$\downarrow$             \\ \midrule
\multicolumn{3}{c}{($\flow^\nn$ directly as output without $\refnet$)}                    & 3.14 / 4.59                   \\
\checkmark        &                     &              & 3.09 / 3.91                   \\
                  & \checkmark          &              & 3.47 / 4.37                   \\
                  & \checkmark          & \checkmark   & 3.48 / 4.26                   \\
\checkmark        &                     & \checkmark   & \textbf{3.04} / 3.83                   \\
\checkmark        & \checkmark          &              & \textbf{3.04} / 3.81          \\
\checkmark        & \checkmark          & \checkmark   & \textbf{3.04} / \textbf{3.73} \\ \bottomrule
\end{tabular}
\vspace{-1em}
\end{table}

%% file: tables/general.tex
\begin{table}[!t]
\setlength{\tabcolsep}{0.5em}
\caption{\textbf{Performance of generalization} on different splits of the animals subset in \ddtfd{} dataset. Checkmarks indicate inclusion for training. \epe{} and AccS are measured on full point clouds w/o synchronization.}
\label{tbl:general}
\vspace{-1em}
\footnotesize
\centering
\begin{tabular}{@{}ccc|cc|cc@{}}
\toprule
\multicolumn{3}{c|}{Training set} & \multicolumn{4}{c}{Test on $(\Cset_2, \Aset_3)$} \\ \midrule
$(\Cset_1, \Aset_1)$   & $(\Cset_2, \Aset_2)$   & $(\Cset_\human, \Aset_\human)$   &  \epe{} $\downarrow$  &  $\Delta$ $\downarrow$  & AccS $\uparrow$ & $\Delta$ $\uparrow$   \\ \midrule
\checkmark & \checkmark &            &  3.16    &  0.0\%   &    63.5    &  0.0  \\
\checkmark &            &            &  3.32    &  +5.06\% &    61.9    &  -1.6  \\
           &            & \checkmark &  10.22   &  +223\%  &    34.3    &  -29.2  \\ \bottomrule
\end{tabular}
\end{table}

%% file: tables/worobust.tex
\begin{table}[!t]
\setlength{\tabcolsep}{5.0pt}
\caption{\textbf{Comparison of different synchronization methods} on \dsapien~dataset. `Fixed Iter.' denotes optimizing for a fixed number of iterations, while `Converged Iter.' means to iterate until convergence. `Time' measures the full run time of the synchronization module.}
\label{tbl:worobust}
\centering
\vspace{-1em}
\footnotesize
\begin{tabular}{@{}l|c|ccc@{}}
\toprule
\multirow{2}{*}{}      & \multirow{2}{*}{\begin{tabular}[c]{@{}c@{}}Fixed Iter. \\ \epe~$\downarrow$\end{tabular}} & \multicolumn{3}{c}{Converged Iter.} \\
                       &                                                                                      & \epe~$\downarrow$         & Time (s)   & \# Iter (k)  \\ \midrule
Init. \wosync              & \multicolumn{2}{c}{3.48 $\pm$ 2.41}                                                                      & -          & -           \\ \midrule
Spectral~\cite{huang2021multibodysync}    & \multicolumn{2}{c}{3.24 $\pm$ 1.57}                                                                      & -          & -           \\
w/o $\rho(\cdot)$        & 3.31 $\pm$ 1.90                                                                          & 3.18 $\pm$ 1.79   & 4.2        & 0.2         \\
w/o Basis Precond. & 3.17 $\pm$ 1.87                                                                          & 3.03 $\pm$ 1.57   & 16.7       & 0.72        \\
w/o Alter. Optim.  & 3.27 $\pm$ 1.87                                                                          & \textbf{3.00} $\pm$ \textbf{1.55}   & (45.5)       & 1.0        \\
Ours Full              & \textbf{3.09} $\pm$ \textbf{1.68}                                                                          & 3.03 $\pm$ 1.56   & 7.0        & 0.32        \\ \bottomrule
\end{tabular}
\end{table}

%% file: sections/grp5-ending.tex
\section{Conclusion}
\label{sec:conclude}


In this paper we present \name, a novel method for multiway non-rigid point cloud registration. 
In our framework, we learn the basis functions defined on the domain of point clouds and represent the registration under the framework of functional maps, endowing us with both efficiency and flexibility.
The maps are then digested by our robust synchronization module to incorporate cycle-consistency constraints into the solution, producing more accurate scene flow estimates.
A potential application of non-rigid reconstruction is shown in \cref{fig:slav}.
Our method leverages ideas from computer graphics community concerned with processing clean geometry and applies these ideas to the task of computer vision where occlusions and noise are inevitable.
The good performance of \name{} on a full spectrum of cases gives hope for future works in bridging the gap between these two schools.

\parahead{Limitations}
Despite the state-of-the-art performance, \name~ also has some limitations which we point out in the following and illustrate in \cref{fig:failure}:
(1) the estimated bases sometimes overly-regularize the motion under large displacements and topological changes, causing wrong parts to be stitched together;
(2) \name~is less robust to handling motion ambiguities caused by severe occlusions --- in the extreme case $\C$ will be ill-conditioned if the corresponding functional coordinates amount to zero; and
(3) we find that the trained network has difficulties to generalize to new domains (\eg from humans to animals/objects). This is partially due to the global field-of-view we employed in our basis network being easily fitted to the entire object structure.
\rev{(4) Due to the large and possibly ambiguous solution space of non-rigid registration, we limit all the datasets to a relatively small view change, in line with the weakly-supervised setting in \cite{sharma2020weakly}.
Future studies on transformation in-/equi-variant backbones may help relax such an assumption.}
\rev{(5) Performance-wise, the synchronization module requires $O(K^2)$ pairwise maps as input and the computational cost grows quadratically \wrt the number of views. The root cause of this is the fully-connected graph $\mathcal{G}$ we have assumed: A principled way to build and solve on a sparsified graph $\mathcal{G}$, or to use batch-wise or incremental synchronization algorithms are good directions to take.}

\parahead{Future Work} Besides addressing aforementioned limitations, \name~leaves ample room for future works, including better representation of the deformation flow and incorporating more priors to capture complicated motions.

%% file: sections/grp6-appendix.tex
\section{Self-supervised Flow Loss}
\label{apsec:selfsuploss}

In this section we detail the self-supervised flow loss used to train our networks, including the following three terms:

\begin{itemize}[topsep=0pt,leftmargin=*]
\setlength{\itemsep}{3pt}
\setlength{\parskip}{0pt}
\item \textbf{Chamfer Loss} encourages $\X_k$ to move as close as possible to $\X_l$:
\begin{align}
    \loss_\mathrm{chamfer} := & \sum_{ \x_k^w \in \X_k^w } \min_{ \x_l \in \X_l } \lVert \x_k^w - \x_l \rVert_2^2 \hspace{0.5em} + \\ & \sum_{\x_l \in \X_l} \min_{\x_k^w \in \X_k^w} \lVert \x_k^w - \x_l \rVert_2^2,
\end{align}
where $\X_k^w := \X_k + \flow_{kl}$ and the elements $\x$ in $\X$ refer to the points in the point set.
\item \textbf{Smoothness Loss} encourages the local smoothness of the predicted scene flow for nearby points:
\begin{equation}
    \loss_\mathrm{smooth} := \sum_{\x_{i} \in \X_k} \frac{1}{|N(\x_{i})|} \sum_{\x_{j} \in N(\x_{i})} \lVert \flow_{k,i} - \flow_{k,j} \rVert_2^2,
\end{equation}
where $(\cdot)_i$ and $(\cdot)_j$ are the point indices and $N(\x)$ is the set of $k$ nearest neighbors of $\x$.
\item \textbf{Laplacian Loss} enforces the geometric structure depicted by the laplacian coordinate vector is preserved after warping, giving:
\begin{equation}
    \loss_\mathrm{lap} := \sum_{\x_k^w \in \X_k^w} \lVert \Delta(\x_k^w) - \Delta(\x_l^\mathrm{inter}) \rVert_2^2,
\end{equation}
where $\x_l^\mathrm{inter}$ is the interpolated position of $\x_k^w$ on $\X_l$ and the laplace operator is defined as:
\begin{equation}
    \Delta(\x_i) := \frac{1}{|N(\x_i)|} \sum_{\x_j \in N(\x_i)} (\x_j - \x_i).
\end{equation}
\end{itemize}

For more details please refer to \cite{wu2019pointpwc}. The only difference between the loss we used and theirs is that we remove the multi-scale supervision on each hierarchy.

\section{Solving the Synchronization Problem}
\label{apsec:eigen}

We use an alternating method to solve~\cref{eq:sync} as described in the main text. The optimization starts with the pairwise initialization $\{\C_{kl}^0 \}$ obtained from \cref{eq:amin} and iterates over the following two steps:
\begin{enumerate}[leftmargin=*]
\setlength{\itemsep}{0pt}
\setlength{\parskip}{0pt}
    \item \textbf{Optimizing $\{\Hfunc_k\}$ with fixed $\{\C_{kl}\}$.} As demonstrated in \cite{huang2021multibodysync}, after fixing $\{\C_{kl}\}$, the problem becomes a generalized Rayleigh problem with the orthonormality constraint. The spectral solution $\Hfunc := [ \Hfunc_1^\top, \dots, \Hfunc_K^\top ]^\top \in \R^{KM \times V}$ are the $V$ eigenvectors of the graph connection Laplacian matrix $\Lap \in \R^{KM \times KM}$ corresponding to the $V$ smallest eigenvalues. The matrix $\Lap$ is constructed as follows:
    \begin{align}
    \label{eq:wgcl}
    \hspace{-3mm}\Lap = \begin{bmatrix}
    \Lap_1 & -\hat{\C}_{12} & \dots & -\hat{\C}_{1K} \\
    -\hat{\C}_{21} & \Lap_2 & \dots & -\hat{\C}_{2K} \\
    \vdots & \vdots & \ddots & \vdots \\
    -\hat{\C}_{K1} & -\hat{\C}_{K2} & \dots & \Lap_K,
    \end{bmatrix},
    \end{align}
    where
    \begin{equation}
        \begin{aligned}
            \hat{\C}_{kl} & := \C_{kl} + \C_{lk}^\top, \\
            \Lap_{k} & := \sum_{(k,l)\in \Edge} \Id_M + \sum_{(l,k)\in \Edge} \C_{lk}^\top \C_{lk}.
        \end{aligned} 
    \end{equation}
    \item \textbf{Optimizing $\{\C_{kl}\}$ with fixed $\{\Hfunc_k\}$.} With $\{\Hfunc_k\}$ being considered as constant, the full problem is then decomposed into several small independent quadratic problems optimizing each $\C_{kl}, (k,l) \in \Edge$, which is tackled using the robust IRLS technique shown in \cref{algo:irls}. During each inner iteration the optimal solution $\C_{kl}^{[t]}$ (with $t$ being the iteration count) is given by solving a closed-form $M^2 \times M^2$ linear system:
\begin{equation}
\label{eq:c}
\begin{aligned}
\vec(\C_{kl}^{[t]}) = & \hspace{0.3em} \mathbf{\Gamma}_{kl}^{-1} \bb_{kl}, \\ 
\mathbf{\Gamma}_{kl} := & \hspace{0.3em} \Id \otimes \A_{kl}^\top \A_{kl} + \Hfunc_l \Hfunc_l^\top \otimes \Id, \\
\bb_{kl} := & \hspace{0.3em} \vec(\A_{kl}^\top \B_{kl} + \Hfunc_k \Hfunc_l^\top),
\end{aligned}
\end{equation}
    where $\otimes$ is kronecker product, $\vec(\cdot)$ is column-wise vectorization, $\A_{kl}:=\hwt_{kl} \basis_k^{(kl)}$, $\B_{kl}:=\hwt_{kl} \basis_l^{(kl)}$ and the computation of the diagonal weight matrix $\hwt_{kl} \in \R^{I_{kl} \times I_{kl}}$ is described in \cref{apsec:irls}.
\end{enumerate}
The convergence criteria of the above optimization is defined as the mean relative change of each element in all pairwise $\C_{kl}$ being less than $3\times10^{-4}$ or exceeding the maximum number of iterations, which we set to 20.



\section{Details of the IRLS Solver}
\label{apsec:irls}

The solution to the robust argmin problem is used in both the end-to-end network training (\cref{eq:amin}) and the synchronization module (\cref{eq:sync}). Hence the computation must be fast and differentiable (\ie $\frac{\partial E_{kl}}{\partial \C_{kl}^0}$ for back propagation).
We use the IRLS algorithm shown in \cref{algo:irls} for the forward pass and the backward pass is achieved by auto-differentiation via unrolling the iterations.
The optional input of $\C_{kl}^{[0]}$ is provided as the result from the last iteration (or simply $\C_{kl}^0$ at the beginning of the optimization) while solving the synchronization problem.

\begin{algorithm2e} [h!]
\DontPrintSemicolon
\SetKwInOut{Input}{Input}\Input{Aligned bases $\basis_k^{kl}, \basis_l^{kl}$ and optional initial $\C_{kl}^{[0]}$.}
\SetKwInOut{Output}{Output}\Output{Optimal $\C_{kl}$.}
\For{$t\gets 1$ \KwTo $T$}{
    \uIf{$t = 1$ {\normalfont \textbf{and}} $\C_{kl}^{[0]}$ not exists }{
        $\hwt_{kl} \gets \Id_{I_{kl}}$.
    }
    \Else{
        $r_i \gets \lVert \basis_{l,i:}^{(kl)} - \basis_{k,i:}^{(kl)} \C^{[t-1]}_{kl} \rVert, \hspace{0.5em} i \in [1, I_{kl}] $, \\ \vspace{0.5em}
        $\hwt_{kl} \gets \diag \big( \omega(r_{1}), ..., \omega(r_{I_{kl}}) \big)^{\frac{1}{2}} $.
    }
    $\A_{kl}\gets\hwt_{kl} \basis_k^{(kl)}, \hspace{0.5em} \B_{kl}\gets\hwt_{kl} \basis_l^{(kl)}$, \\  \vspace{0.5em}
    Set $\C_{kl}^{[t]} \gets \A_{kl}^+ \B_{kl} $ \emph{or} apply \cref{eq:c} for solving \cref{eq:amin} \emph{or} \cref{eq:sync}, respectively.
}
$\C_{kl} \gets \C_{kl}^{[T]}$.
\caption{Solving the argmin problem with IRLS.}
\label{algo:irls}
\end{algorithm2e}

During the weight computation of $\hwt_{kl}$, we define $\omega(\cdot)$ as the influence function of the Huber robust kernel~\cite{huber1992robust}, formulated as:
\begin{equation}
    \omega(r) := \begin{cases}
        1 & \text{if $|r| < \kappa$}\\
        \kappa / |r| & \text{if $|r| \geq \kappa$}
      \end{cases},
\end{equation}
where the scale factor $\kappa$ is empirically chosen as 0.05.
With such a scale, not only the robustness is maintained but the resulting energy landscape also has few local minima, hence making the convergence reasonably fast within a few iterations.
For efficiency considerations, we let $T=2$ for solving \cref{eq:amin} and $T=1$ for solving one iteration of \cref{eq:sync}.
Note that although we iterate only once for the synchronization, \cref{algo:irls} still needs to be run multiple times during the alternating optimization scheme of $\{\Hfunc_k\}$ and $\{\C_{kl}\}$.

\begin{figure}[!t]
\centering
\includegraphics[width=\linewidth]{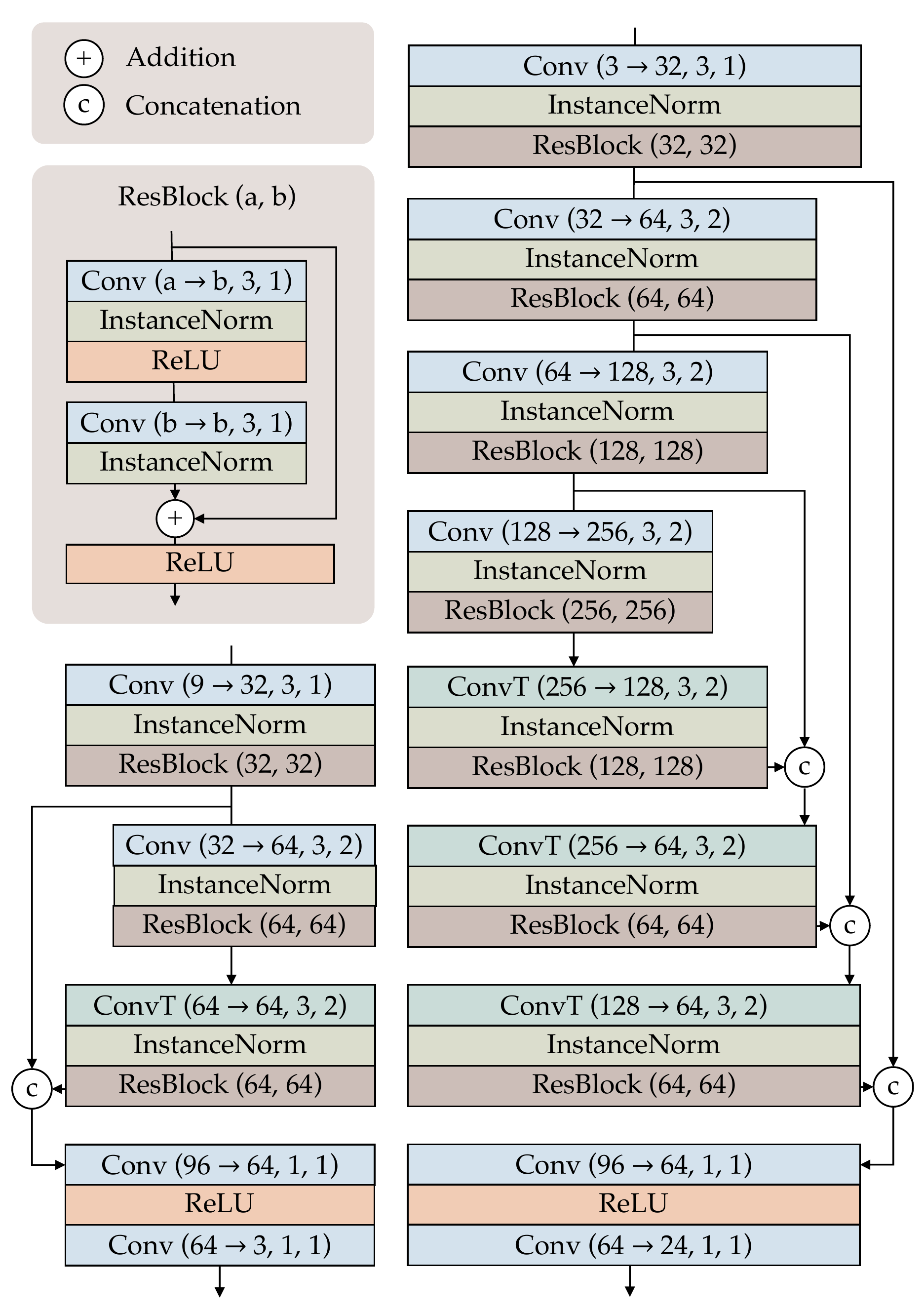}
\vspace{-1.5em}
\caption{\textbf{Network architectures.} Top-left: legends and the definition of the residual block (ResBlock). Bottom-left: illustration of the 2-layer $\refnet$. Right: illustration of the 4-layer $\basisnet$. The same structure is applied in $\descnet$, and only the number of channels change from (32, 64, 128, 256) to (32, 96, 64, 192) as described in the main text.}
\vspace{-1em}
\label{fig:backbone}
\end{figure}

\section{\rev{Network Architectures}}
\label{apsec:network}

The architectures of all three networks ($\basisnet$, $\descnet$ and $\refnet$) used in our experiments are adopted from the work of \cite{choy2020deep}, and implemented using MinkowskiEngine~\cite{choy20194d}.
As 3D point cloud data are sparsely distributed in space, running a 3D convolution on a dense voxel grid scattered from the sparse points usually introduces large memory consumption and waste of computation over empty regions.
To this end, a `sparse' version of convolution is introduced in \cite{choy20194d} that create and track voxels only on non-empty regions, maintained using a hash table to enable fast neighborhood queries.
In practice, such sparse operators are not only significantly faster and more memory-friendly, but also capable of offering impressive performance for many downstream tasks due to the convolutional nature.

In \name{}, we use the \texttt{ResUNet} structure, a U-Net with skip connections and residual blocks within each hierarchy, as illustrated in \cref{fig:backbone}.
The layers and their arguments are defined as follows:
\begin{itemize}[leftmargin=*]
    \item Conv ($a \rightarrow b$, $K$, $S$): Sparse convolution with input channels $a$, output channels $b$, kernel size $K$ and stride $S$. Note $S=2$ means spatially shrinking the feature map sizes by two.
    \item InstanceNorm and ReLU are instance normalization~\cite{ulyanov2016instance} and rectified linear unit~\cite{nair2010rectified} respectively.
    \item ConvT ($a \rightarrow b$, $K$, $S$): Sparse \emph{transposed} convolution with input channels $a$, output channels $b$, kernel size $K$ and stride $S$. Here $S=2$ means spatially dilating the feature map sizes by two.
\end{itemize}

When creating the sparse voxels for convolution, input points have to be discretized.
We use $0.01^3$ voxel size for all datasets except for LidarKITTI, where we use $0.1^3$.
As conversions back and forth between point and voxel representations are inefficient and possibly harm performance, in practice we directly work on the level of \emph{voxels}, including the putative correspondences $\corr$ and the soft permutation matrices $\Perm_{kl}$.
Final per-point flow vectors are interpolated from nearby voxels via (subscript ${kl}$ omitted):
\begin{equation}
    \flow_{i:} := \frac{\sum_{j\in \knn_i} \flow_{j:}^v \lVert \x_i - \x_j^v \rVert_2^{-1} }{\sum_{j\in \knn_i} \lVert \x_i - \x_j^v \rVert_2^{-1}},
\end{equation}
where $\knn_i$ returns the indices of nearby voxels to point $\x_i \in \X_i$, and $\x^v$, $\flow_{j:}^v$ are coordinates and flow vectors on the center of the $j$-th voxel, respectively.

Still, we choose to keep the descriptions in the main text based on \emph{points} for brevity and clarity, and not to break the generality of our method in the sense that the backbones can be replaced with more advanced ones free of such voxelizations.

\section{\rev{Miscellaneous Details}}
\label{apsec:misc}

\parahead{EmbAlign~\cite{marin2020correspondence} Baseline}
The original version of EmbAlign uses a simple PointNet backbone to extract the features. 
To enable a fair comparison, we attempt to replace the PointNet with the sparse-conv backbone used in our method. 
However, we found that replacing both the networks (\ie, the `embedding network $\mathcal{N}$' and the second (probe) network `$\mathcal{G}$') leads to bad convergence.
By exhaustively trying out different combinations, we find that only replacing the second (probe) network `$\mathcal{G}$' leads to the best result.
This is probably due to the design choice in EmbAlign, that requires the probe functions to be explicitly projected onto the bases, preventing it from recover high-frequency features that are beneficial for accurate correspondences.

\parahead{DeepShells~\cite{eisenberger2020deep} Baseline}
For DeepShells, the use of normal data makes the method unstable because accurate estimation of normal (especially the orientation) from point cloud is non-trivial. 
In the final version we use for testing, we hence remove the normal term from the iterative optimization algorithm (Smooth Shell).
We also tried to replace the SHOT feature with FPFH but the results are worse than directly feeding the absolute coordinates into the network.

\parahead{Frame Indices for \cref{fig:dd}}
The frame indices we used for the qualitative results shown in \cref{fig:dd} are respectively: 
\ding{172} Seq008: 400, \underline{500}, 600, 800;
\ding{173} Seq009: 200, \underline{400}, 600, 800;
\ding{174} Seq012: 300, \underline{600}, 800, 900;
\ding{175} Seq004: 600, 800, \underline{900}, 1000;
\ding{176} Seq026: \underline{200}, 300, 500, 600;
\ding{177} Seq027: 300, 400, 600, \underline{700}.
Underlined frames are shown as reference (\texttt{ref}) to which other frames are warped.

\parahead{Categories of \dsapien{}}
The dataset is generated including the following categories:
box, dishwasher, display, door, eyeglasses, faucet, kettle, knife, laptop, lighter, microwave, oven, phone, pliers, refrigerator, safe, scissors, stapler, storageFurniture, toilet, trashCan, washingMachine, and window.

\parahead{Split of the Dataset in \cref{subsec:ablation}(\ref{para:general})}
The splits we use are:
$\Cset_1$: bear, bucks, bull, cat, crocodile, dino, doggie, dragon, elephant, elk, fox, goat, hippo, hog, huskydog, lioness, moose, pig, puma, seabird, sheep, zebra;
$\Cset_2$: bunny, canie, duck, grizz, leopard, milkcow, procy, rhino, tiger;
Validation set: cattle, cetacea, chicken, deer, rabbit, raccoon, raven.